\documentclass{ws-jcsc}
\usepackage{verbatim}
\usepackage{enumerate,graphics,subfigure,epsf,graphicx,epsfig,color}
\usepackage{multirow}%
\usepackage{latexsym,amsmath,amssymb,amsfonts,mathrsfs}
\usepackage{mathrsfs}%
\usepackage[title]{appendix}%
\usepackage{xcolor}%
\usepackage{textcomp}%
\usepackage{manyfoot}%
\usepackage{booktabs}%
\usepackage{algorithm,multirow,colortbl}
\usepackage{algorithmicx}%
\usepackage{algpseudocode}%
\usepackage{listings}%
\usepackage{hyperref}%
\newcommand{\R}{{\mathbb R}}

\newcommand{\be}{\begin{eqnarray}}
\newcommand{\ben}{\begin{eqnarray*}}
\newcommand{\en}{\end{eqnarray}}
\newcommand{\enn}{\end{eqnarray*}}

\DeclareMathOperator*{\argmax}{arg\,max}
\DeclareMathOperator*{\argmin}{arg\,min}

\graphicspath{{figures/}}
\begin{document}


%
\catchline{}{}{}{}{}
%

\title{Careful seeding for the $k$-medoids algorithm with incremental $k$-means++ cluster construction}

\author{Difei Cheng\footnote{First Author}}

\address{Institute of Automation, Chinese Academy of Sciences,\\
Beijing, 100190, China\\
chengdifei@amss.ac.cn}

\author{Yunfeng Zhang}

\address{Academy of Mathematics and Systems Science, Chinese Academy of Sciences, \\
Beijing ,100190, China\\
zhangyunfeng@amss.ac.cn}

\author{Ruinan Jin\footnote{Corresponding Author}}

\address{School of data science, The Chinese university of Hong Kong, Shenzhen,\\
Shenzhen, 518172, China\\
jinruinan@cuhk.edu.cn}

\maketitle


\begin{abstract}
The $k$-medoids algorithm is a popular variant of the $k$-means algorithm and widely used in pattern
recognition and machine learning.
A main drawback of the $k$-medoids algorithm is that an improper initialization can cause it to get trapped in a local optima.
An incremental $k$-medoids algorithm named improved $k$-medoids algorithm (INCKM) was recently proposed to overcome this drawback which firstly combines the incremental initialization algorithm with the $k$-medoids algorithm. The INCKM algorithm constructs a subset of candidate medoids with a parameter choosing procedure for initialization, but it always fails when dealing with imbalanced datasets.
In this paper, we propose a novel incremental $k$-medoids algorithm called incremental $k$-means$++$ algorithm (INCKPP) which dynamically increases
the number of clusters from $2$ to $k$ through a nonparametric and stochastic $k$-means$++$ search procedure.
The INCKPP algorithm overcomes the parameter selection problem in the INCKM algorithm,
improves the clustering performance, and can deal with imbalanced datasets well.
However, the INCKPP algorithm is not computationally efficient enough. To deal with this,
we further propose a fast INCKPP algorithm (called INCKPP${}_{\textbf{sample}}$) which can improve the computational
efficiency while maintaining the clustering performance.
Extensive experiments on both synthetic and real-world datasets, including imbalanced datasets, illustrate that the proposed algorithm outperforms than other compared algorithms.
\end{abstract}



\section{Introduction}\label{sec:intro}

Clustering is an important class of unsupervised learning methods. In recent decades, hundreds of clustering algorithms have been proposed \cite{cao2023clustering,wu2008top,schubert2017dbscan,cheng2023fast,li2018discriminatively} and have been widely used in many applications such as data mining \cite{sato2019data}, vector quantization \cite{guo2020accelerating},
dimension reduction \cite{boutsidis2015randomized}, and manifold learning \cite{CanasPR12}.
The aim of a clustering task \cite{kaufman2009finding} 
is to partition a set of objects into clusters such that the objects within a cluster have a high degree
of similarity but are dissimilar to objects in other clusters.

The $k$-means algorithm \cite{wu2008top} is one of the most popular clustering algorithms,
which minimizes the sum of the squared distance from each data point to its nearest center.
It is popular for its easy implementation, linear time complexity, and convergence guarantee \cite{selim1984k}. 
However, the $k$-means algorithm is sensitive to outliers.
The $k$-medoids (or the Partition Around Medoids (PAM)) algorithm \cite{rdusseeun1987clustering} is a variant 
of the $k$-means algorithm, which minimizes the sum of the dissimilarity between each data point and
the center it belongs to. Compared to the $k$-means algorithm, k-medoids algorithm overcomes the problem of sensitivity to outliers in $k$-means algorithm, and is more flexible in choosing dissimilarity measures to deal with diverse data distributions. 


Since 2003, in order to further improve the clustering performance, several incremental initialization algorithms \cite{likas2003global,bagirov2008modified,lai2010fast,bai2013fast,bagirov2011fast} have been proposed to find a global minimizer for the $k$-means-like algorithms such as the $k$-means and the $k$-medoids algorithms.
These incremental initialization algorithms which are different from the traditional initialization algorithm \cite{park2009simple,arthur2007k,rodriguez2014clustering,erisoglu2011new,tan2023statistical,vzalik2008efficient,broin2015alignment,xie2016k,zadegan2013ranked,dinh2020k} take an incremental approach to clustering that dynamically adds
one cluster center at a time through a deterministic global search procedure consisting of $n$
executions of the $k$-means algorithm from suitable initial positions, where $n$ is the size of the data set.
These algorithms improve the clustering performance significantly, but introduce a much higher computational complexity,
compared to the linear computational complexity in the traditional initialization algorithms such as $k$-means$++$.

An incremental $k$-medoids algorithm named improved $k$-medoids algorithm (INCKM) was proposed \cite{yu2018improved}, which first combines the incremental initialization algorithm with the $k$-medoids algorithm for initialization to achieve a better clustering 
performance than many commonly used algorithms 
\cite{park2009simple,rodriguez2014clustering}. 
But INCKM needs to determine a subset of candidate medoids subset with a parameter choosing procedure during initialization. Although INCKM gives a wide range of choices for the parameters, it often fails to deal with some complex datasets and, in particular, imbalanced datasets. Therefore, we propose a novel incremental $k$-medoids algorithm (called INCKPP) which proceeds in an incremental manner and dynamically increases the number of clusters and centers from $2$ to $k$ through a non-parametric, stochastic $k$-means$++$ search procedure. 
The main contributions of this paper are summarized as follows.
\begin{itemize}
	\item INCKPP addresses the difficulty in selecting parameters in the INCKM algorithm \cite{yu2018improved}, while improves the clustering performance and can also deal with imbalanced data sets very well.
	\item INCKPP is inefficient computationally. To address this issue, we propose an improved INCKPP algorithm named INCKPP${}_{sample}$ which is faster than INCKPP, while keeping the clustering performance of INCKPP algorithm. Extensive experiments have been conducted on 14 synthetic datasets and seven real datasets to illustrate that our algorithm outperforms than other compared most commonly used algorithms such as INCKM \cite{yu2018improved}, FKM \cite{park2009simple} and KPP \cite{arthur2007k} algorithms.
	
\end{itemize}

The remaining part of the paper is organized as follows. Section \ref{sec:related-works} introduces
the related work. Section \ref{sec:Inckpp} gives details of the proposed algorithms.
Extensive experimental results are conducted in Section \ref{sec:Exp} on synthetic and real-world datasets
in comparison with INCKM, FKM and the $k$-means$++$ algorithm (KPP). 
Conclusions are given in \ref{sec:conclusion}.

\section{Related work}\label{sec:related-works}

\subsection{Simple and fast $k$-medoids algorithm (FKM)}

Given a data set $\textbf{X}=\{x_i\in\R^p|i=1,2,\ldots,n\}$, the simple and fast $k$-medoids (FKM)
algorithm in \cite{park2009simple} aims to minimize the Sum of the Errors (SE) between $x_i\in\textbf{X}$
and the cluster center $c_j\in\textbf{C}=\{c_j\in\textbf{X}|j=1,2,\ldots,K\}$:
\begin{equation}\label{eq:se}
	\text{SE}(\textbf{C})=\sum_{i=1}^{n}\sum_{j=1}^{k}s_{ij}\,d(x_i,c_j),
\end{equation}
where $\textbf{S}=\{s_{ij}|i=1,2,\ldots,m; j=1,2,\ldots,k\}$ is the assignment index set with
$s_{ij}=1$ indicating that $x_i$ is assigned to the $j$th cluster and $s_{ij}=0$ otherwise,
$d(x_i,c_j)$ is the distance between $x_i$ and $c_j$.
FKM algorithm selects the cluster centers, which are called \textit{medoids}, from the data set and
updates the new medoids of each cluster which is the data point minimizing the sum of its distances to
other data points in the cluster.


\subsection{$k$-means++}

The $k$-means++ algorithm in \cite{arthur2007k} is an augmented $k$-means algorithm which uses a simple,
randomized seeding technique in the $k$-means algorithm.
It chooses the first center randomly with equal probability and sequentially selects the rest
centers $x\in X$ with the probability ${d(\textbf{x})^2}/{\sum_{\textbf{x}'}d(\textbf{x}')^2}$, where $d(\textbf{x})$ denotes
the minimum distance from $\textbf{x}$ to the closest center that we have already chosen.
The $k$-means++ algorithm can exclude the influence of the outliers.

\subsection{The improved $k$-medoids algorithm}\label{subsec:Inckm}

The improved $k$-medoids algorithm (INCKM) \cite{yu2018improved} is an incremental algorithm for
solving the $j$-medoids problem from $j=2$ to $j=k$ in turn.
It is easy to find the optimal solution of the $1$-medoids problem, which is the data point minimizing
the total distance from the data point to other data points in the data set.
For solving the $(j+1)$-medoids problem, the solution of the $j$-medoids is used as the first $j$
initial cluster centers and one tries to find the $(j+1)$ initial cluster centers.
The INCKM algorithm first computes the variances of the data set as follows:
\be\label{eq:var}
\sigma &=& \sqrt{\frac{1}{n-1}\sum^n_{i=1}dist(x_i,\bar{x})^2},\\ \label{eq:varr}
\sigma_{i} &=& \sqrt{\frac{1}{n-1}\sum^n_{j=1}dist(x_i,x_j)^2},
\en
where $\bar{x}=\sum_{i=1}^nx_i/n$ and $dist(x_i,x_j)$ is the distance between $x_i$ and $x_j$.
It then constructs a candidate set $S=\{x_i|\sigma_i\le\lambda\sigma,\;i=1,\ldots,n\}$ with
the $(j+1)$ initial centers given by
\be\label{eq:inck}
c_j=\argmax_{x_i\in S}\{d_i|i=1,\ldots,n\},
\en
where $d_i$ is the distance between $x_i$ and the closest center we have already chosen
and $\lambda$ is a stretch factor which needs to be given in advance.
The stretch factor in INCKM was recommended to be chosen
between $1.5$ and $2.5$. However, this parameter range may certainly not be suitable for all types
of data distributions and, in particular, for some imbalanced data distributions.


\section{The proposed incremental $k$-medoids algorithm}\label{sec:Inckpp}

In this section, we propose a novel incremental $k$-medoids algorithm, called incremental $k$-means$++$ algorithm (INCKPP), which proceeds in an incremental manner, attempting to optimally add one new cluster center at each stage through $k$-means++ searching. More specifically, it solves the clustering problem with k clusters as follows. INCKPP first initializes the first cluster center at the position of the point with the smallest sum of distances to the other points, which is the position of the center of the solution of the clustering problem with one cluster.  The INCKPP then proceeds to solve the clustering problem with two clusters. In this stage, the first cluster center initialized at the position of cluster center in the solution of the clustering problem with one center and chooses the second cluster center $x\in X$ corresponds to the probability ${d(\textbf{x})^2}/{\sum_{\textbf{x}'}d(\textbf{x}')^2}$, where $d(\textbf{x})$ denotes
the minimum distance from $\textbf{x}$ to the first cluster center, then executes the classical FKM algorithm to get the result. More generally,  we assume that the solution of the clustering problem with t-1 cluster centers has been obtained by INCKPP, in order to solve the clustering problem with t cluster centers, the first $t-1$ cluster centers initialized at the position of the centers in the solution for the clustering problem with t-1 centers, while choses the $t$-th cluster center $x\in X$ corresponds to the probability ${d(\textbf{x})^2}/{\sum_{\textbf{x}'}d(\textbf{x}')^2}$, where $d(\textbf{x})$ denotes
the minimum distance from $\textbf{x}$ to the first $t-1$ cluster center chosen already. 
Being different from the INCKM algorithm, INCKPP has no parameter to be given in advance except for the cluster
number $k$ and can deal with complex and imbalanced data distributions very well,
as demonstrated in experiments. Further, INCKPP has the ability to exclude the influence of the outliers
in view of using $k$-means++ searching in execution.
More details of the INCKPP algorithm is given in Algorithm \ref{alg:Inckpp} below.

\begin{algorithm}[!htb]
	\caption{INCKPP}\label{alg:Inckpp}
	\begin{algorithmic}[1]
		\Require dissimilarity matrix $\textbf{D}$, data set $X$ and cluster number $k$
		\Ensure cluster center $\textbf{C}=\{c_1,\ldots,c_k\}$
		\State Initialize $C=\{c_1\}$:
		$$c_1=\argmin_{x_i\in X}\sum_j d(x_i,x_j)$$.
		\vspace{-0.4cm}
		\For {j=2,...,k}
		\State Find $c_j$ through $k$-means++ searching, that is, choosing $c_j$ with the probability
		${d(\textbf{x})^2}/({\sum_{\textbf{x}'}d(\textbf{x}')^2})$.
		\State Update $C=C\cup{c_j}$.
		\State Updata cluster center $C$ by FKM in \cite{park2009simple}.
		\EndFor
		\Return $C$
	\end{algorithmic}
\end{algorithm}
Regardless of the complexity in calculating the distance matrix in all k-medoids-like methods, the complexity of updating the medoids in the incremental methods such as INCKPP and INCKM is $O(tn\sum_{j=2}^{k})=O(tnK^2)$. Although the iteration number $t$ of INCKPP algorithm is smaller than that of INCKM algorithm as demonstrated in experiments, the complexity of its update medoids is still greater than that of FKM algorithm, which is $O(tnk)$.
To improve the time efficiency of INCKPP we further propose a sampled INCKPP algorithm
called INCKPP$_{sample}$. In INCKPP$_{sample}$, INCKPP is executed
first on randomly sampled $p$ percent of each data set once as a pre-search procedure. Generally, we recommend values of $p$ ranging from  5 to 15, and the value of $p$ is robust as demonstrated in experiments as follows.
Then, by using the result of the pre-search procedure as the initial centers, INCKPP$_{sample}$
executes FKM on the whole dataset to find the final centers. The complexity of updating medoids  in NCKPP$_{sample}$ is $O(t_{1}n_{1}k^2+t_{2}nk)$, where $n_{1}\ll n$ and the number of  iteration $t_1,t_2$ is smaller than that of INCKM and FKM as shown in experiments results.

\section{Experiments}\label{sec:Exp}

In this section, we conduct a number of experiments to evaluate the clustering performance of
the proposed INCKPP and INCKPP$_{sample}$ algorithms.
All the experiments are conducted on a single PC with Intel Core 2.6GHz i5 CPU (2 Cores) and 8G RAM.

\subsection{The compared algorithms}\label{subsec:compare}

In the experiments, we first compare INCKPP with INCKM \cite{yu2018improved}
and then compare INCKPP$_{sample}$ with KPP \cite{arthur2007k}, KPP$_{sample}$, FKM \cite{park2009simple}
and FKM$_{sample}$, where
\begin{itemize}
	\item KPP is the algorithm which chooses $k$ initial medoids through $k$-means++ searching and then uses
	FKM as a local search procedure,
	\item KPP$_{sample}$ is the algorithm which first runs KPP on randomly sampled $p$ percent of each
	data set as a pre-search procedure and then runs FKM with the result in the pre-search procedure
	as initial centers to find the final result,
	\item FKM$_{sample}$ is the algorithm which first runs FKM on randomly sampled $p$ of each data set
	($p=10$, i.e., $10$ percent of the data points are used) as a pre-search procedure and then runs
	FKM with the result in the pre-search procedure as initial centers to find the final result.
\end{itemize}

\subsection{The data sets}\label{subsec:data}

\subsubsection{Overview of the data sets}\label{subsub:overview}

The synthetic datasets can be download from the clustering datasets
website\footnote{\url{http://cs.joensuu.fi/sipu/datasets/}\label{fn:1}}.
The imbalance data set contains $6500$ two-dimensional data points. The experiments consider its three
subsets that contain the classes $\{6,7\}$ (denoted as imbalance$_2$), $\{2,6,7,8\}$ (denoted as imbalance$_4$),
$\{1,3,4,5,6,8\}$ (denoted as imbalance$_6$) and the whole imbalance data set (denoted as imbalance).
The datasets in Dim-sets contains $9$ well separated Gaussian clusters with the dimension varying from $2$ to $15$.
The datasets in S-sets contains $15$ Gaussian clusters with different degrees of clusters overlapped.
We choose two data sets from Shape-sets which consist of hyper-spherical clusters.
The real world data sets are from UCI Machine Learning Repository\footnote{\url{http://archive.ics.uci.edu/ml/}\label{fn:2}} . Handwritten digits data set
contains $7494$ sixteen-dimensional data points. In the experiments, we consider three subsets that
contain the class $\{1,3,5\}$ (denoted as pendigits${}_3$), $\{0,2,4,6,7\}$ (denoted as pendigits${}_5$),
$\{0,1,2,3,4,5,6,7\}$ (denoted as pendigits${}_8$) and the whole handwritten digits data set (denoted as pendigits).
In the several real datasets we chosen, the class number varies from $2$ to $10$, the dimension of the
data points varies from $3$ to $16$ and the size of the data sets varies from $215$ to $7494$.
The Euclidean distance is used in both synthetic and real datasets although other dissimilarity can
also be used in the $k$-medoids type algorithms.

\begin{table}[!thb]
	\centering
	\caption{Overview of the synthetic datasets and the real datasets.}
	\begin{tabular}{cccccc}
		\toprule
		&\multicolumn{2}{c}{data set} & attribute & class & size\\
		\midrule
		\multirow{14}{*}{Synthetic datasets}
		&\multirow{4}{*}{Imbalanced-sets}
		&imbalance${}_2$ & 2 & 2 & 2100\\
    	&&imbalance${}_4$ & 2 & 4 & 4200\\
		&&imbalance${}_6$ & 2 & 6 & 6300\\
		&&imbalance & 2 & 8 & 6500\\
		\cmidrule{2-6}
		&\multirow{4}{*}{Dim-sets}
		&dim${}_2$ & 2 & 9 & 1351\\
		&&dim${}_6$ & 6 & 9 & 4051\\
		&&dim${}_{10}$ & 10 & 9 & 6751\\
		&&dim${}_{15}$ & 15 & 9 & 10126\\
		\cmidrule{2-6}
		&\multirow{4}{*}{S-sets} 
		& S${}_1$ & 2 & 15 & 5000\\
		&& S${}_2$ & 2 & 15 & 5000\\
		&& S${}_3$ & 2 & 15 &5000\\
		&& S${}_4$ & 2 & 15 &5000\\
		\cmidrule{2-6}
		&\multirow{2}{*}{Shape-sets}
		& R15 & 2 & 15 & 600\\
		&& D31 & 2 & 31 &3100\\
		\cmidrule{2-6}
		\multirow{7}{*}{Real datasets}
		&\multirow{4}{*}{Handwritten digits-sets}
		&pendigits${}_3$&16&3&2218\\
		&&pendigits${}_5$&16&5&3838\\
		&&pendigits${}_8$&16&8&6056\\
		&&pendigits&16&10&7494\\
		\cmidrule{2-6}
		&\multirow{4}{*}{other real datasets}
		&newthyroid&5&3&215\\
		&&banknote&4&2&1372\\
		&&yeast&8&10&1484\\
	
		\botrule
	\end{tabular}
\end{table}

\subsubsection{Attribute normalization} \label{subsec:Atribute}

The normalization
\begin{equation}\label{eq:min_max}
	x_{normalization}=\frac{x-x_{min}}{x_{max}-x_{min}}
\end{equation}
is used to avoid the attributes with large values dominating the distance calculation and
to get more accurate numerical computation results.

\subsection{Performance criteria} \label{sec:performance}

In the experiments, four criteria are used to evaluate the performance of the algorithm:
1) the average of the sum of errors (aver-SE), 2) the minimum of the sum of errors (min-SE),
3) the number of iterations ($\#iter$) and 4) the number of repeats ($\#repe$) within the CPU time.
In the experiments comparing INCKPP with INCKM,
we take the stretch factor $\lambda=1.4+0.1k$ with $k=1,\ldots,11$ in the INCKM algorithm as
in \cite{yu2018improved}. We regard the value of $\lambda$ with the smallest sum of errors (SE)
as the best parameter and record the corresponding SE which is denoted as min-SE*.
We first run the INCKM algorithm to get an upper bound for the CPU time and record correspondingly
the min-SE*, the best parameter $\lambda$ and $\#iter$ of INCKM.
We then run INCKPP repeatedly as many times as possible within the CPU time of INCKM.
The min-SE, aver-SE, $\#iter$ and $\#repe$ of INCKPP were recorded when time is over.
Similarly, in the experiments comparing INCKPP$_{sample}$ with the four stochastic algorithms, KPP,
KPP$_{sample}$, FKM and FKM$_{sample}$, we first run the INCKPP$_{sample}$ algorithm $N$ times
to get an upper bound of the CPU time used by INCKPP$_{sample}$ and then run KPP, KPP$_{sample}$,
FKM and FKM$_{sample}$ repeatedly as many times as possible within the CPU time used by INCKPP$_{sample}$.
During this process, we recorded min-SE, aver-SE, $\#iter$ and $\#repe$ of the five comparing algorithms.
The values of the CPU time, min-SE, aver-SE, $\#iter$, $\#repe$ are the average of these criteria
over $100$ duplicate tests to reduce the randomness.


\subsection{Experiments on the synthetic datasets} \label{subsec:synthetic}

\begin{figure}[!ht]
	\centering
	\includegraphics[width=0.8\linewidth]{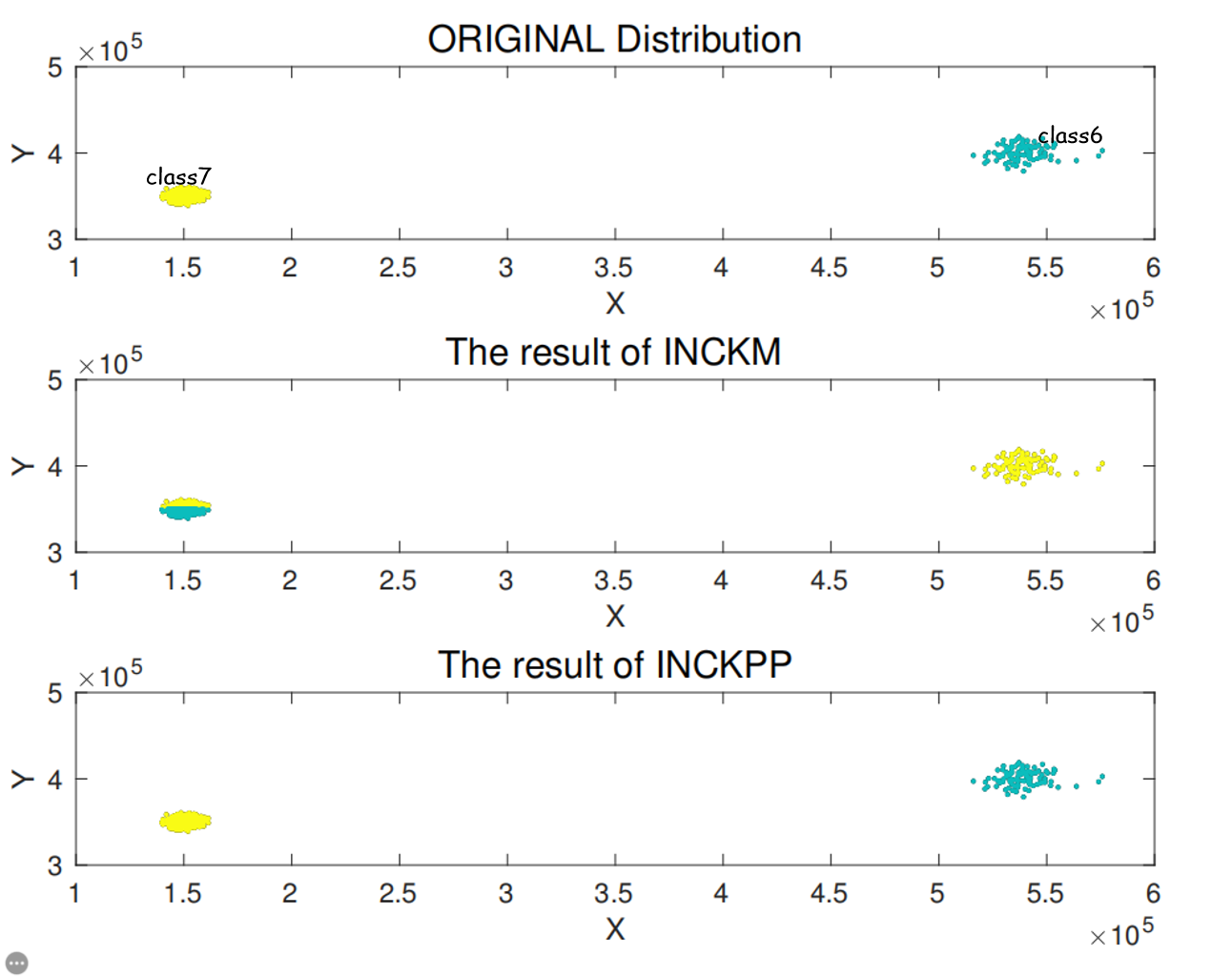}
	\caption{The clustering result on the imbalance${}_{2}$ data set. INCKM gets an incorrect
		clustering result since it initializes two cluster centers in one class, leading to the fact that
		one original class is divided into two classes. Note that INCKPP gets the correct result.
	}\label{fig:imbalance}
\end{figure}

Table \ref{tab:Inckm-Inckpp-synthetic} presents the comparison results of INCKPP with INCKM on the synthetic
data sets. From the results it is seen that the minimum of the sum of errors (min-SE) obtained by INCKPP
is not bigger than the smallest sum of errors (i.e., the min-SE*) obtained by INCKM on $9$ datasets;
in particular, INCKPP achieves a much smaller min-SE than the min-SE* obtained by INCKM on the four
imbalanced datasets with a much smaller number of iterations.
Figure \ref{fig:imbalance} presents the clustering results obtained by INCKM and INCKPP on the imbalance$_2$ data set. For all recommended stretch factors $\lambda$, INCKM fails to correctly initialize
the cluster center, that is, INCKM initializes two cluster centers in one class (Class $7$) which has $20$
times as many samples as another class (Class $6$), so Class $7$ was divided into two classes, as seen
in Figure \ref{fig:imbalance}. However, INCKPP gets the correct clustering result. In fact, for the four imbalanced datasets, INCKM always gets an
incorrect clustering result due to the wrong initialization of INKCM, but INCKPP gets the correct
clustering result.

In what follows, we carry out extensive experiments to compare the fast version of
INCKPP (i.e., INCKPP${}_{sample}$) with KPP$_{sample}$, FKM$_{sample}$, KPP and FKM on synthetic datasets,
where we always denote by $p$ the randomly sampled percentage of the datasets in the pre-search
procedure and by $N$ the running times of INCKPP$_{sample}$.

\begin{table}[!thb]
	\centering
	\caption{The comparisons between INCKM and INCKPP on synthetic datasets. The min-SE is the minimum of SE obtained within the CPU time
		used by INCKM in one run. The $\#repe$ is the number of repeats of INCKPP within the CPU time used by INCKM
		in one run. The $\#iter$ is the number of iterations when the local minimizer is attained.
	}\label{tab:Inckm-Inckpp-synthetic}
		\begin{tabular}{|c|c|c|c|c|c|c|}
			\hline
			\multirow{2}{*}{Data sets}&\multicolumn{3}{c|}{INCKM}&\multicolumn{3}{c|}{INCKPP}\\
			\cmidrule{2-7}
			&$\lambda$&min-SE*&$\#iter$&min-SE&$\#repe$&$\#iter$\\
			\midrule
			imbalance${}_2$&1.5&149.89&9.00&\textbf{75.75}&55.14&\textbf{1.14}\\
			\midrule
			imbalance${}_4$&1.5&132.49&7.00&\textbf{87.29}&21.85&\textbf{1.34}\\
			\midrule
			imbalance${}_6$&1.5&155.75&11.00&\textbf{122.76}&16.51&\textbf{2.92}\\
			\midrule
			imbalance&2.5&140.72&12.45&\textbf{126.92}&18.13&\textbf{3.02}\\
			\midrule
			dim${}_{15}$&1.5&702.09&\textbf{1.00}&\textbf{702.09}&11.85&1.03\\
			\midrule
			dim${}_{10}$&1.5&304.78&1.09&\textbf{304.78}&13.14&\textbf{1.03}\\
			\midrule
			dim${}_{6}$&1.5&131.10&\textbf{1.00}&\textbf{131.10}&11.81&1.03\\
			\midrule
			dim${}_{2}$&1.6&12.62&1.55&\textbf{12.62}&14.34&\textbf{1.04}\\
			\midrule
			S${}_{1}$&1.5&\textbf{181.63}&\textbf{2.00}&183.18&11.16&2.65\\
			\midrule
			S${}_{2}$&2.0&\textbf{219.09}&\textbf{2.82}&221.54&11.38&3.41\\
			\midrule
			S${}_{3}$&1.5&287.32&4.09&\textbf{272.14}&10.45&4.13\\
			\midrule
			S${}_{4}$&1.5&\textbf{253.35}&\textbf{5.36}&255.69&11.71&5.63\\
			\midrule
			D${}_{31}$&1.5&\textbf{109.64}&3.82&115.32&10.19&\textbf{3.30}\\
			\midrule
			R${}_{15}$&1.5&\textbf{16.46}&\textbf{2.00}&16.62&10.83&2.14\\
			\midrule
		\end{tabular}
		
\end{table}

\begin{figure*}[!thb]
	\centering
	\subfigure[imbalance]{\includegraphics[width=0.2\textwidth]{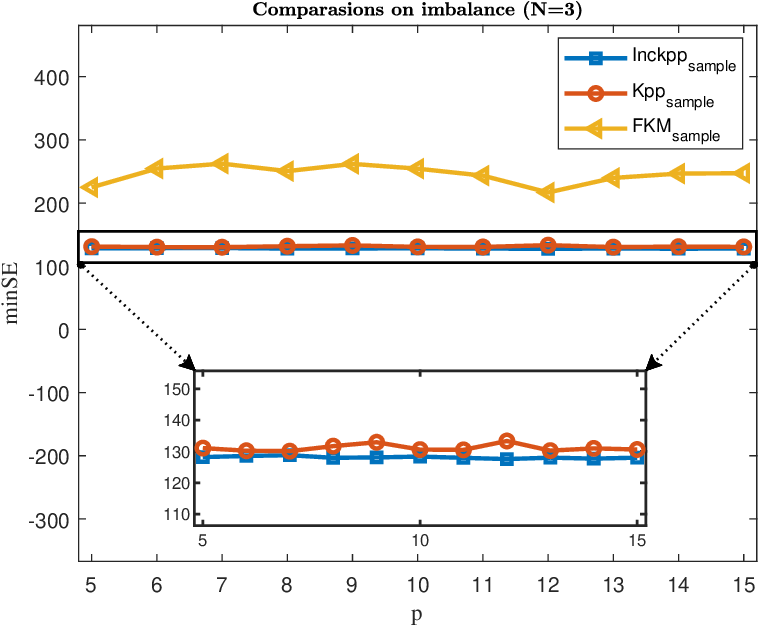}
	}
	\subfigure[imbalance${}_{6}$]{\includegraphics[width=0.2\textwidth]{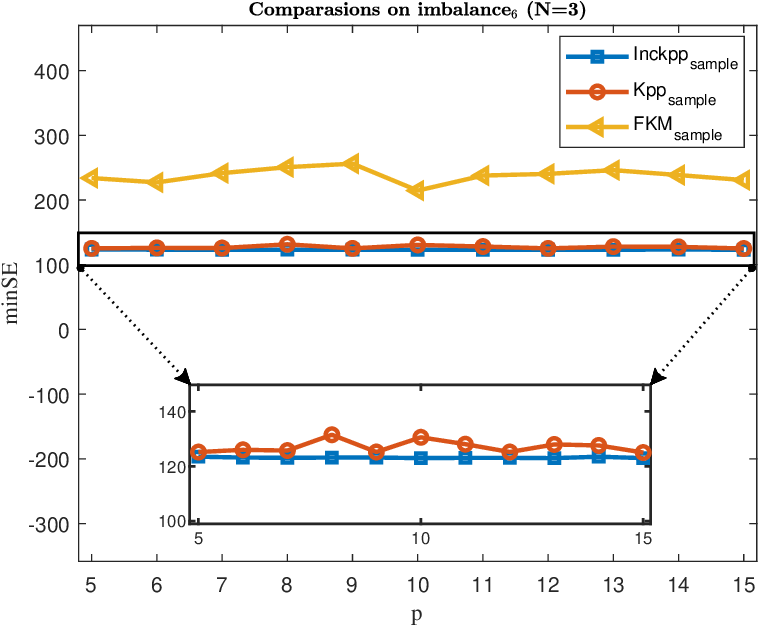}
	}
	\subfigure[imbalance${}_{4}$]{\includegraphics[width=0.2\textwidth]{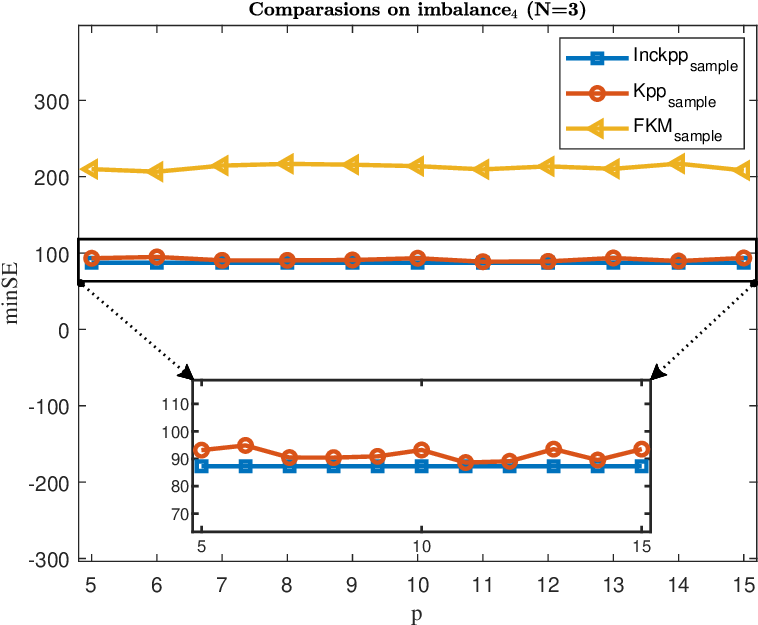}
	}
	\subfigure[imbalance${}_{2}$]{\includegraphics[width=0.2\textwidth]{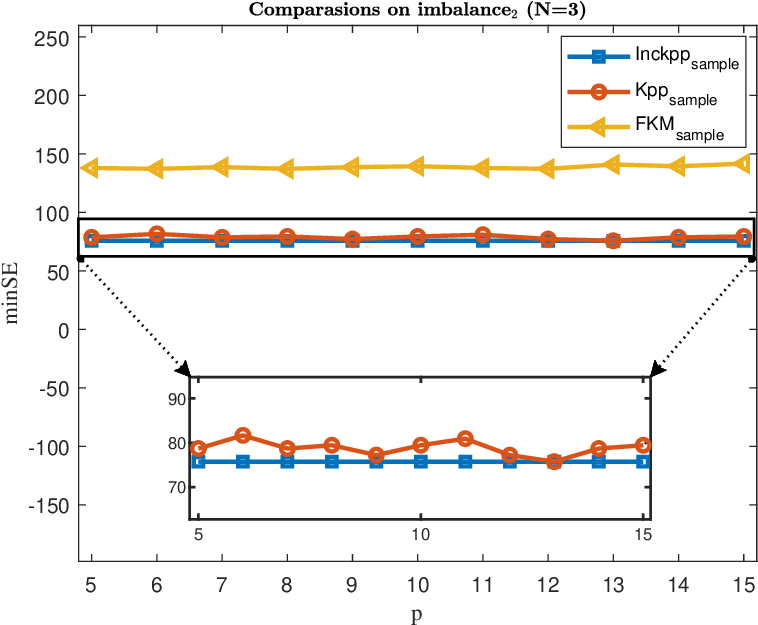}
	}
	\subfigure[S${}_{1}$]{\includegraphics[width=0.2\textwidth]{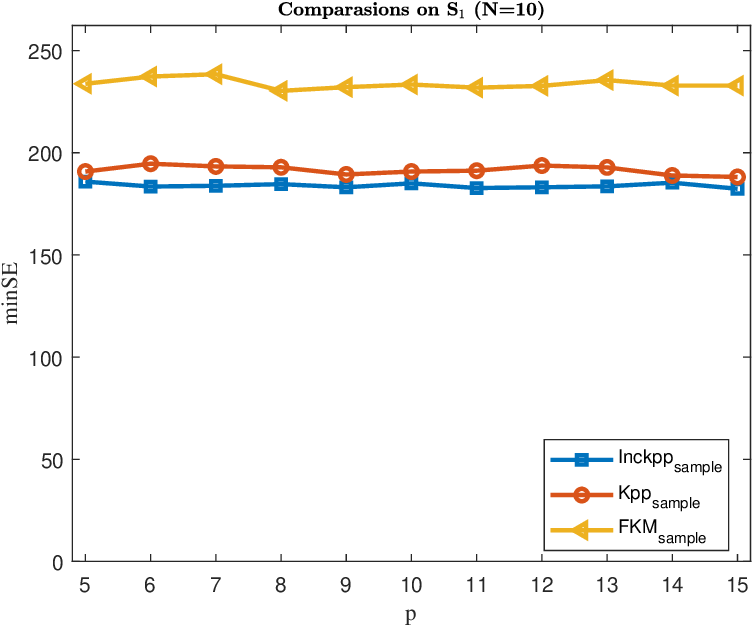}
	}
	\subfigure[S${}_{2}$]{\includegraphics[width=0.2\textwidth]{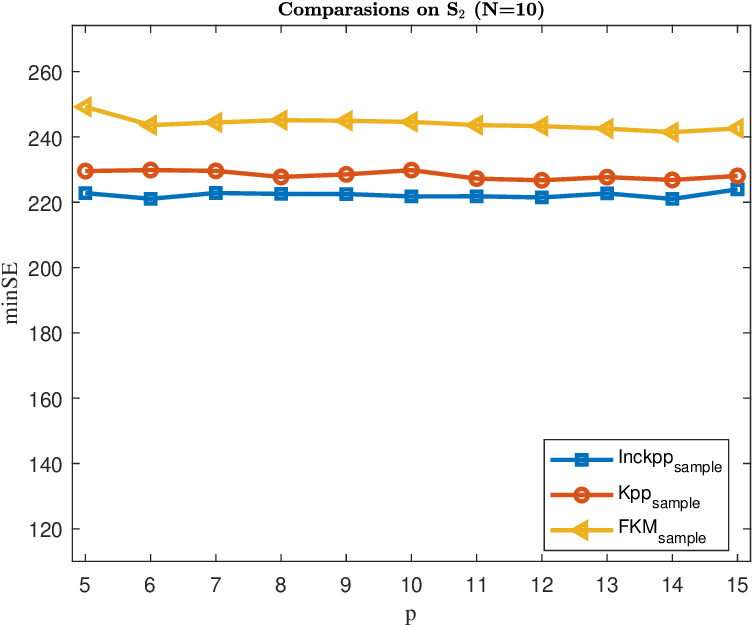}
	}
	\subfigure[S${}_{3}$]{\includegraphics[width=0.2\textwidth]{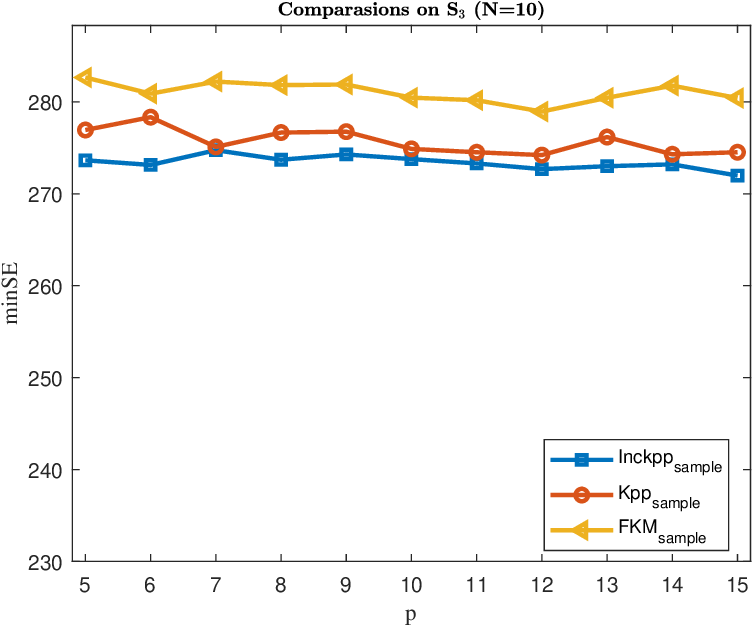}
	}
	\subfigure[S${}_{4}$]{\includegraphics[width=0.2\textwidth]{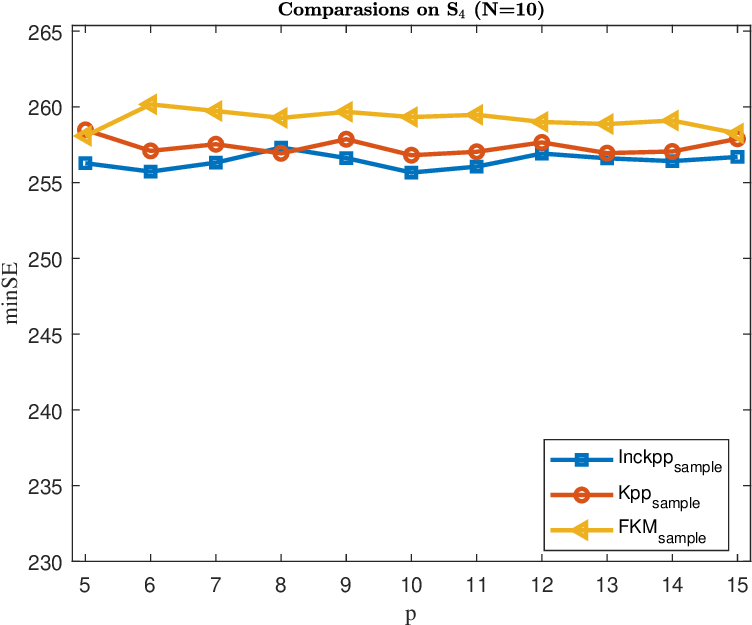}
	}
	\subfigure[dim${}_{2}$]{\includegraphics[width=0.2\textwidth]{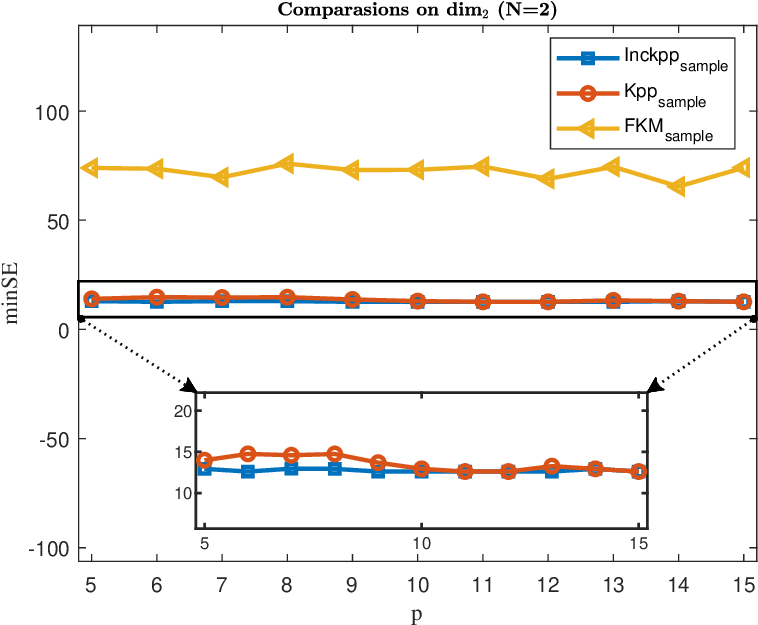}
	}
	\subfigure[dim${}_{6}$]{\includegraphics[width=0.2\textwidth]{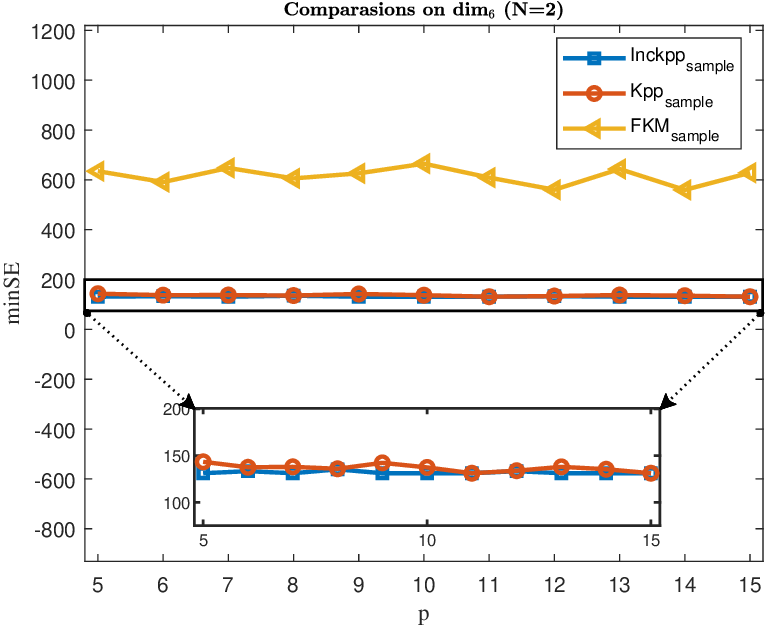}
	}
	\subfigure[dim${}_{10}$]{\includegraphics[width=0.2\textwidth]{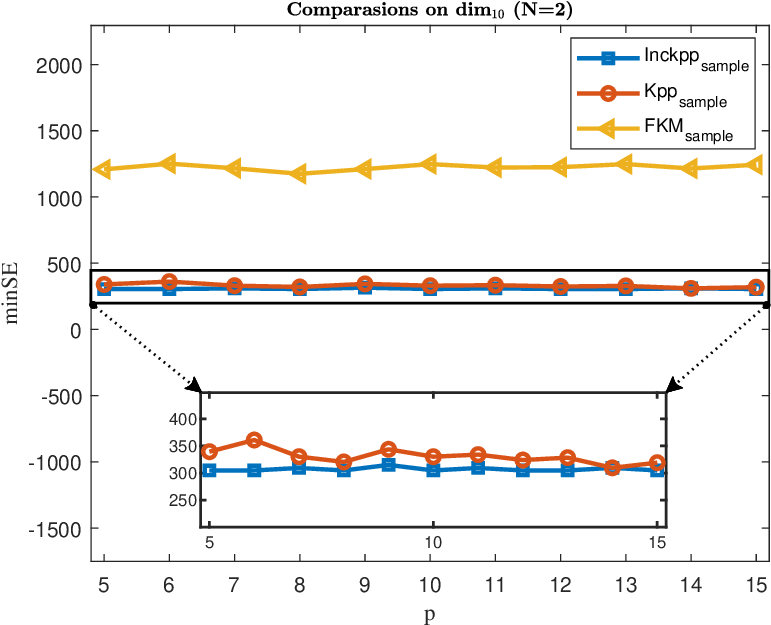}
	}
	\subfigure[dim${}_{15}$]{\includegraphics[width=0.2\textwidth]{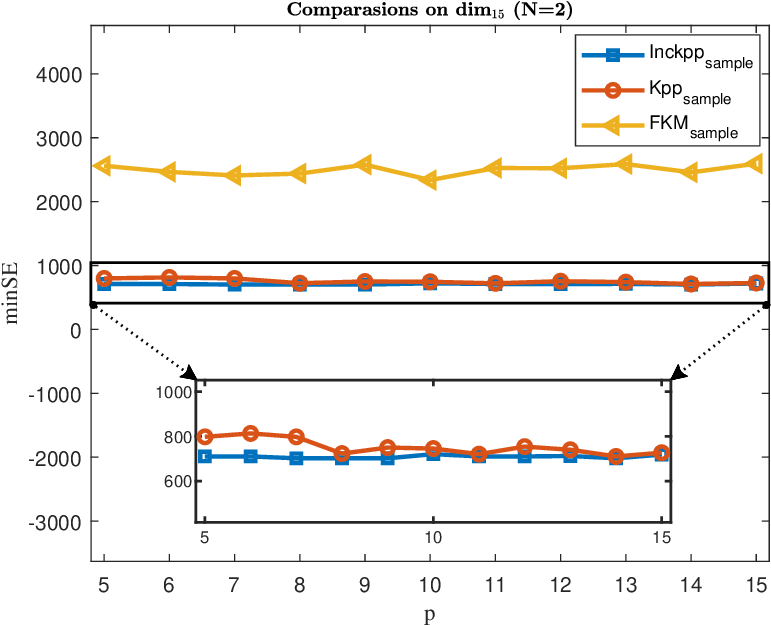}
	}
	\subfigure[R${}_{15}$]{\includegraphics[width=0.2\textwidth]{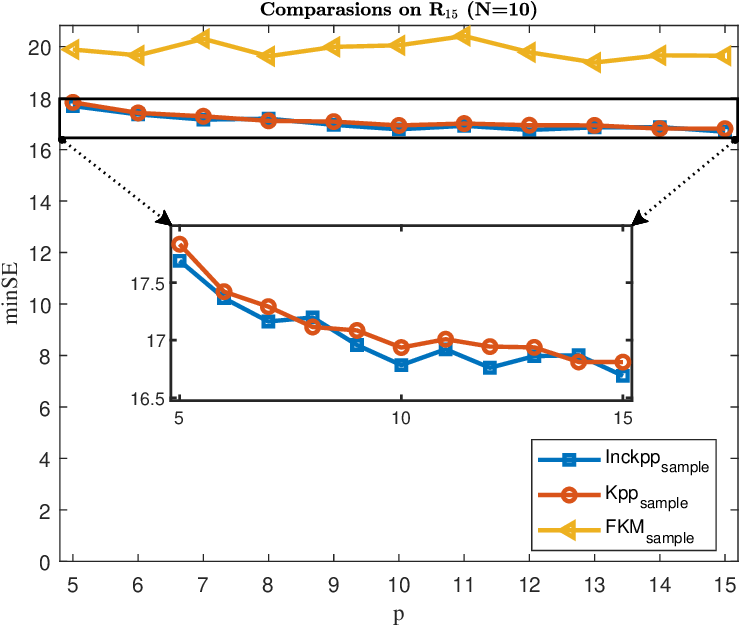}
	}
	\subfigure[D${}_{31}$]{\includegraphics[width=0.2\textwidth]{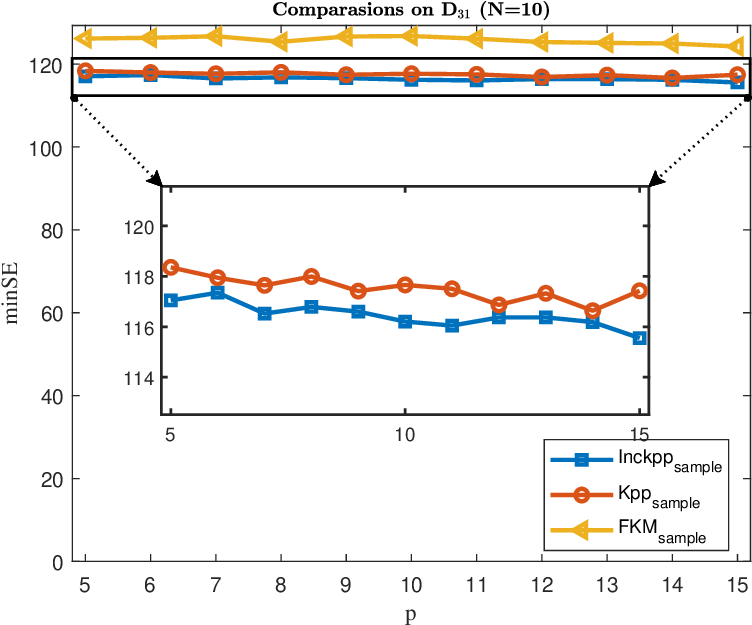}
	}
	\caption{The comparisons on the synthetic datasets with different $p$,
		where $p$ is the randomly sampled percentage of the dataset in the pre-search procedure.
		The values in the figures are the minimum of the sum of errors (min-SE) obtained by each compared
		algorithm within the CPU time used by INCKPP${}_{sample}$ running $N$ times for different $p$.
	}\label{fig:synthetic-p}
\end{figure*}

\begin{figure*}[!thb]
	\centering
	\subfigure[imbalance]{\includegraphics[width=0.2\textwidth]{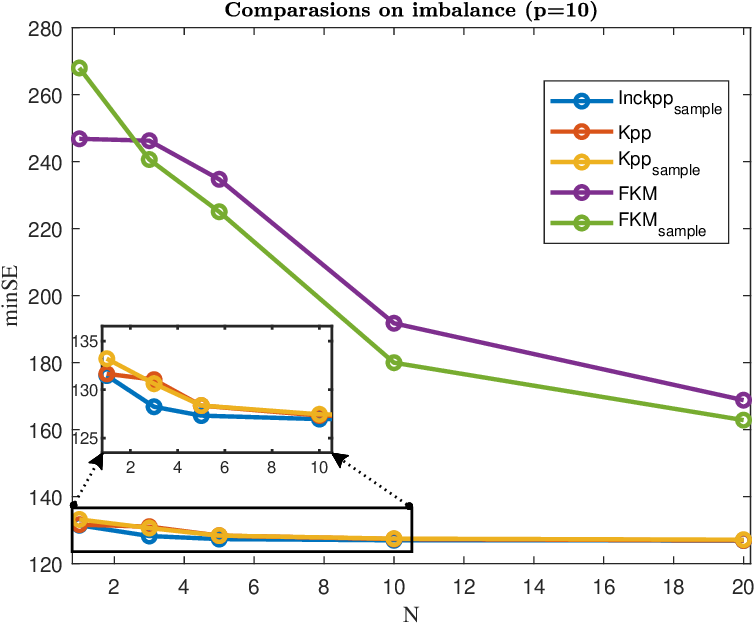}
	}
	\subfigure[imbalance$_6$]{\includegraphics[width=0.2\textwidth]{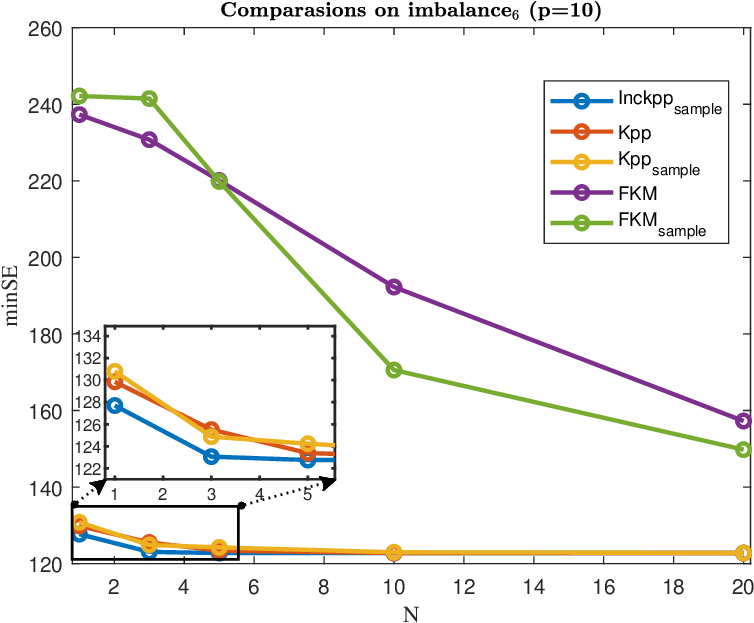}
	}
	\subfigure[imbalance$_4$]{\includegraphics[width=0.2\textwidth]{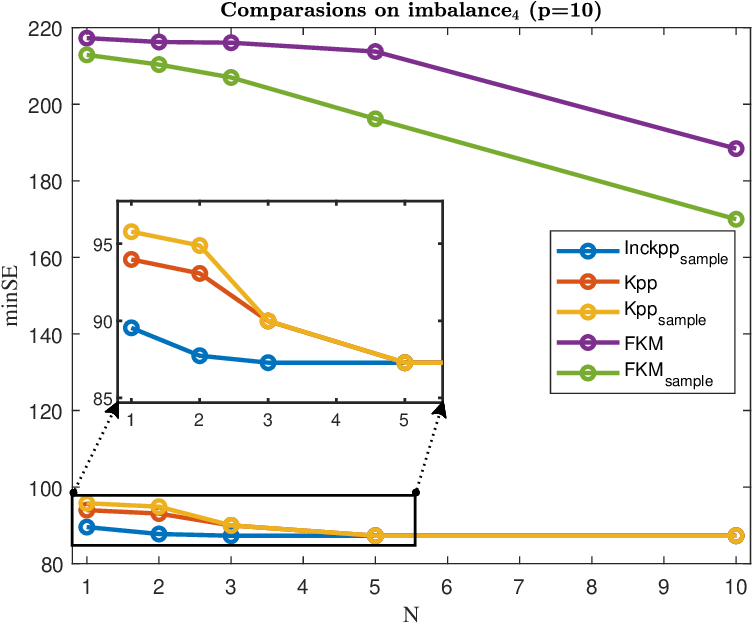}
	}
	\subfigure[imibalance$_{2}$]{\includegraphics[width=0.2\textwidth]{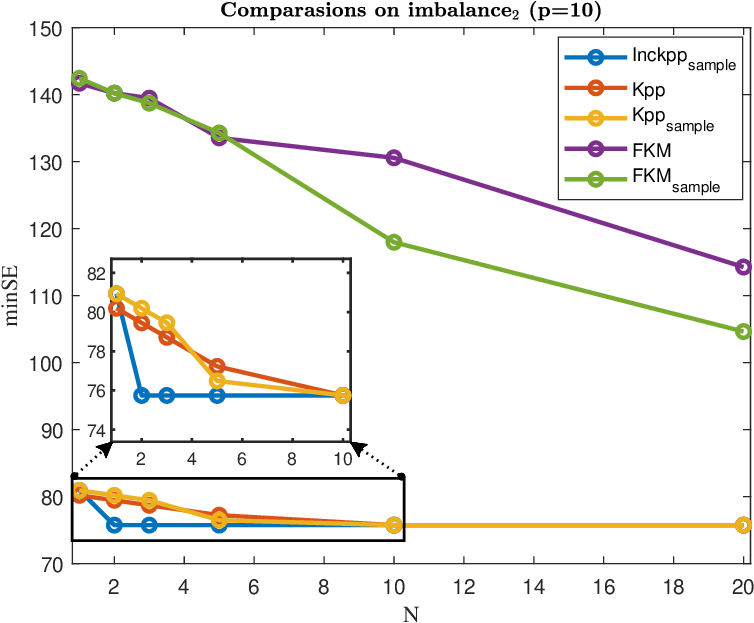}
	}
	\subfigure[S${}_{1}$]{\includegraphics[width=0.2\textwidth]{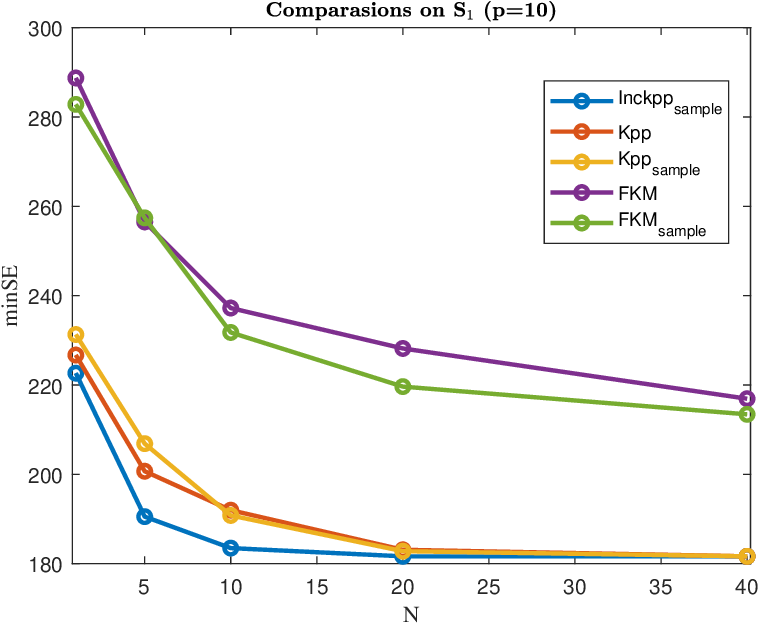}
	}
	\subfigure[S${}_{2}$]{\includegraphics[width=0.2\textwidth]{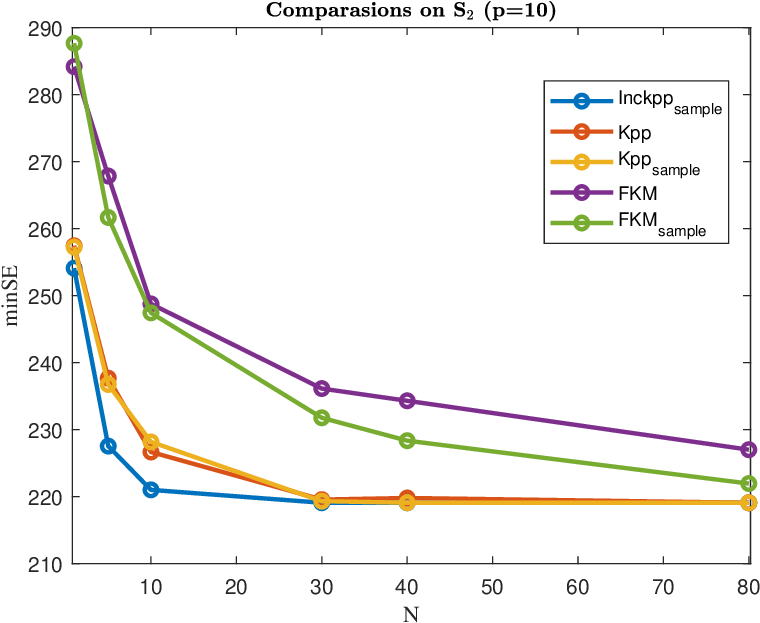}
	}
	\subfigure[S${}_{3}$]{\includegraphics[width=0.2\textwidth]{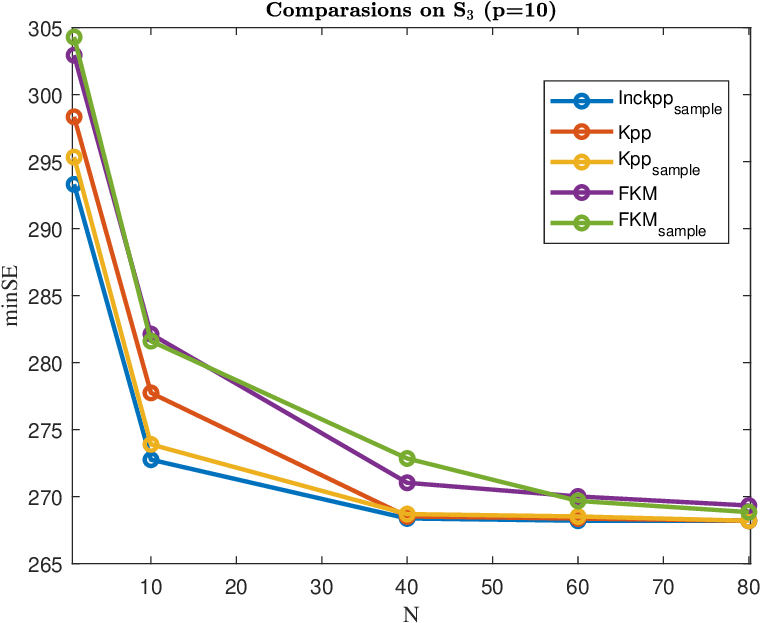}
	}
	\subfigure[S${}_{4}$]{\includegraphics[width=0.2\textwidth]{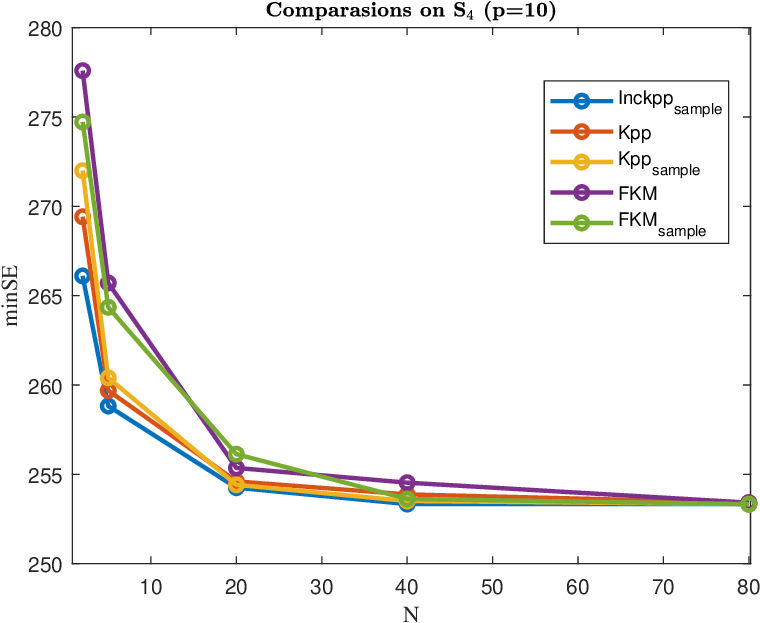}
	}
	\subfigure[dim${}_{2}$]{\includegraphics[width=0.2\textwidth]{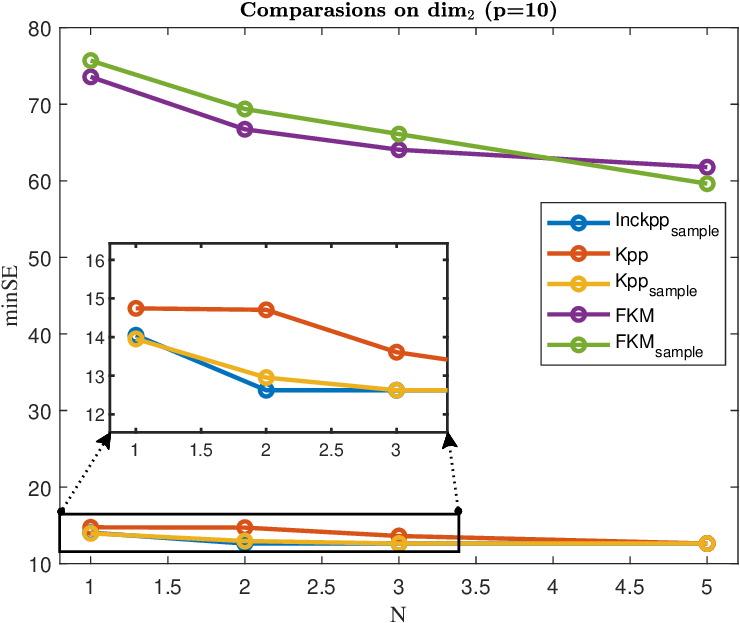}
	}
	\subfigure[dim${}_{6}$]{\includegraphics[width=0.2\textwidth]{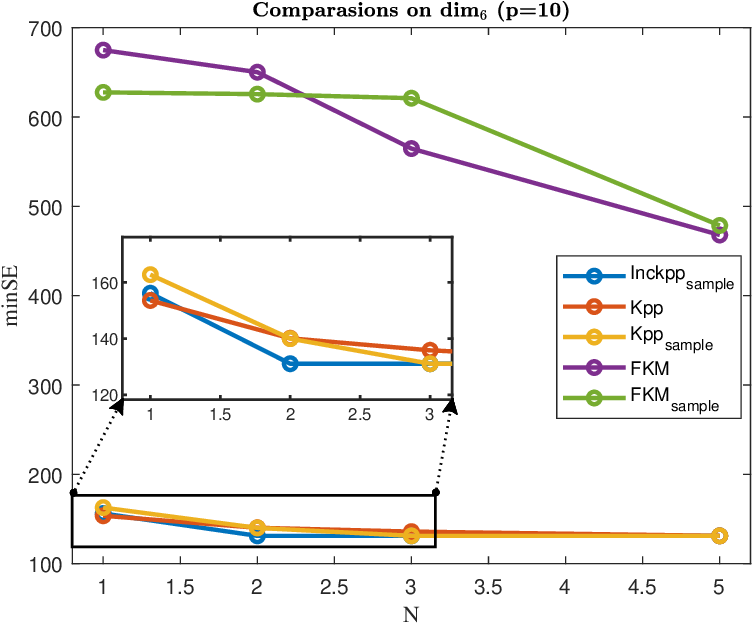}
	}
	\subfigure[dim${}_{10}$]{\includegraphics[width=0.2\textwidth]{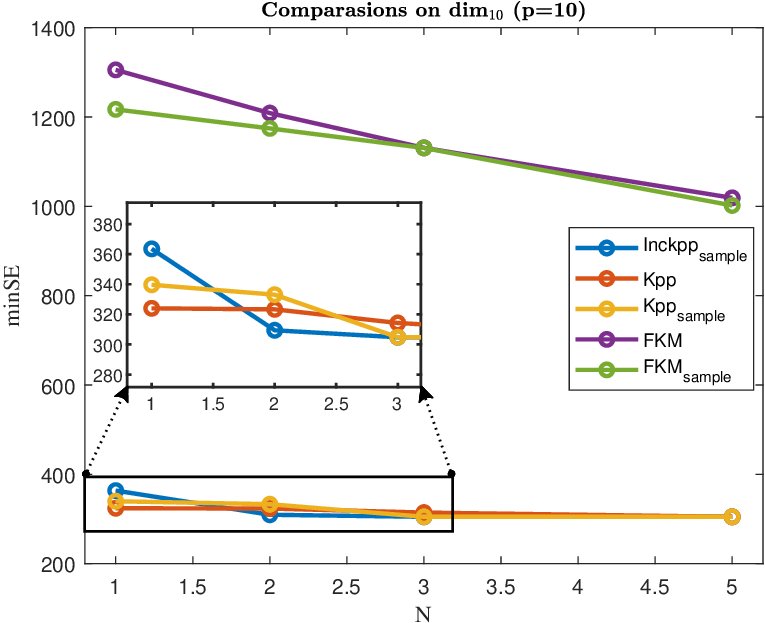}
	}
	\subfigure[dim${}_{15}$]{\includegraphics[width=0.2\textwidth]{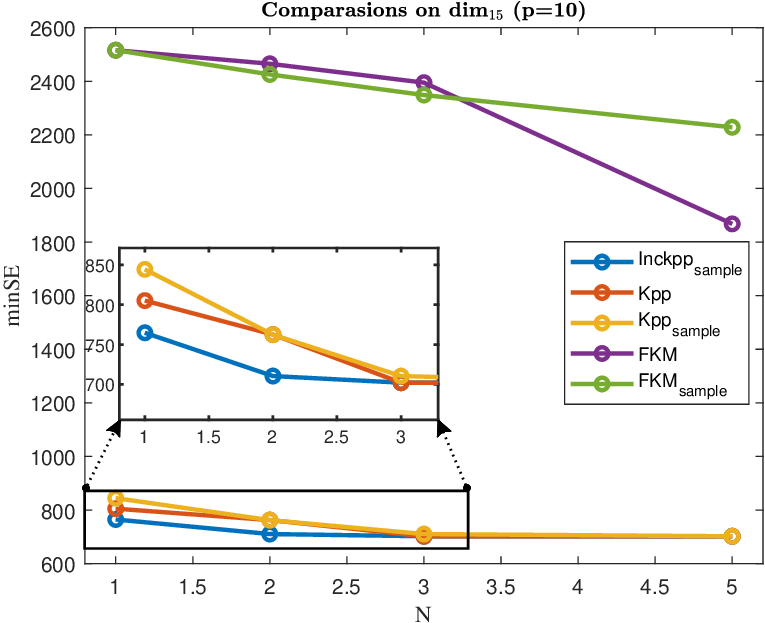}
	}
	\subfigure[R${}_{15}$]{\includegraphics[width=0.2\textwidth]{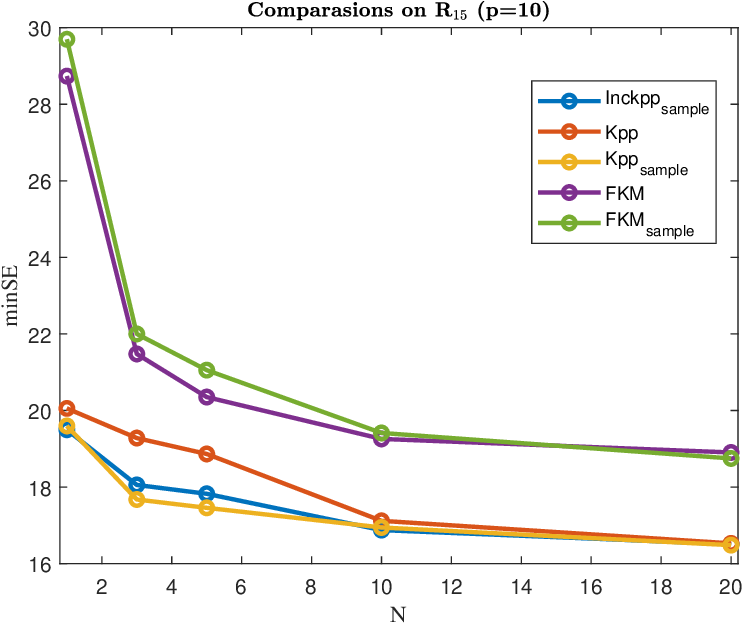}
	}
	\subfigure[D${}_{31}$]{\includegraphics[width=0.2\textwidth]{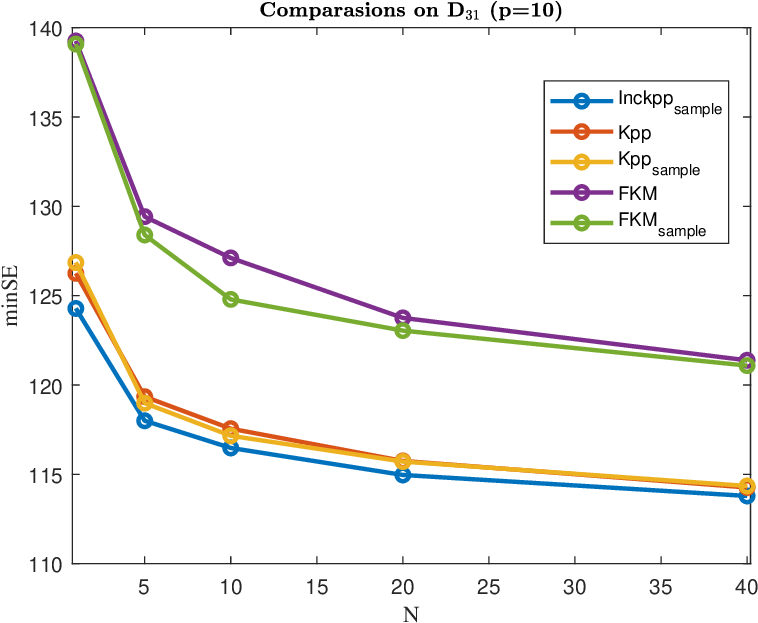}
	}
	
	\caption{The comparisons on the synthetic datasets for different $N$, where $N$ is the running times of INCKPP$_{sample}$ algorithm. The values in the table are the minimum of the sum of errors (min-SE) obtained by each compared algorithm within the CPU time used by INCKPP${}_{sample}$ running $N$ times with $p=10$.
	}\label{fig:synthetic-N}
\end{figure*}

\begin{table*}[!htb]
	\centering
	\caption{The experiment results of INCKPP${}_{sample}$, KPP${}_{sample}$, FKM${}_{sample}$, KPP and FKM
		on the synthetic datasets when $p=10$, where the aver-SE is the average of sum of errors,
		$\#re$ (i.e., $\#repe$) is the number of repetitions within the CPU time used by INCKPP${}_{sample}$
		running $10$ times, and $\#it$ (i.e., $\#iter$) is the number of iterations when the local minimizer
		is achieved.}\label{tab:synthetic}
	\resizebox{\textwidth}{!}{
		\begin{tabular}{|c|c|c|c!{\vrule width 0.8pt}c|c|c!{\vrule width 0.8pt}c|c|c!{\vrule width 0.8pt}c|c|c!
				{\vrule width 0.8pt}c|c|c!{\vrule width 0.8pt}}
			\midrule
			\multirow{2}{*}{Data set}&\multicolumn{3}{c!{\vrule width 0.8pt}}{INCKPP${}_{sample}$}&
			\multicolumn{3}{c!{\vrule width 0.8pt}}{KPP}&\multicolumn{3}{c!{\vrule width 0.8pt}}{KPP${}_{sample}$}
			&\multicolumn{3}{c!{\vrule width 0.8pt}}{FKM}&\multicolumn{3}{c!{\vrule width 0.8pt}}{FKM${}_{sample}$}\\
			\cmidrule{2-16}
			&$\#re$ &$\#it$& aver-SE& $\#re$& $\#it$& aver-SE& $\#re$& $\#it$&aver-SE& $\#re$& $\#it$& aver-SE&
			$\#re$& $\#it$& aver-SE\\
			\midrule
			imbalance&\textbf{10}&\textbf{2.44}&\textbf{132.11}&6.98&4.11&132.84&7.03&4.15&133.05&3.55&10.48&248.72&4.77
			&6.99&246.26\\
			\midrule
			imbalance${}_{6}$&\textbf{10}&\textbf{2.13}&\textbf{127.34}&7.54&3.28&128.99&7.35&3.39&130.03&3.41&8.69
			&240.29&5.09&5.48&239.29\\
			\midrule
			imbalance${}_{4}$&\textbf{10}&\textbf{1.11}&\textbf{90.83}&9.30&1.33&91.99&9.05&1.39&92.76&2.65&7.36
			&212.08&4.24&4.13&212.36\\
			\midrule
			imbalance${}_{2}$&\textbf{10}&\textbf{1.07}&\textbf{78.02}&9.50&1.19&78.65&9.31&1.24&79.50&3.68&5.51
			&142.63&5.38&2.95&142.89\\
			\midrule
			S$_{1}$&\textbf{10}&\textbf{3.17}&\textbf{220.95}&8.96&4.33&230.36&9.14&4.30&230.37&6.38&6.39&287.61
			&7.91&4.81&286.10\\
			\midrule
			S$_{2}$&\textbf{10}&\textbf{4.53}&\textbf{252.35}&8.27&6.48&259.35&8.38&6.50&258.92&5.28&9.00&290.76
			&7.88&6.68&287.63\\
			\midrule
			S$_{3}$&\textbf{10}&\textbf{6.06}&\textbf{293.03}&8.43&8.16&294.77&8.58&8.09&294.95&8.10&9.82&304.50
			&9.46&7.09&304.18\\
			\midrule
			S$_{4}$&\textbf{10}&\textbf{8.21}&\textbf{272.25}&8.60&10.44&273.21&8.64&10.48&273.43&7.18&12.74&278.70
			&8.92&9.95&278.24\\
			\midrule
			dim$_{2}$&10&\textbf{1.07}&\textbf{15.05}&10.84&1.14&15.44&\textbf{11.68}&1.14&15.41&3.84&4.50&73.04
			&4.87&3.62&73.24\\
			\midrule
			dim$_{6}$&10&\textbf{1.02}&\textbf{144.38}&11.01&1.05&150.55&\textbf{11.10}&1.05&152.21&5.36&2.42&642.25
			&7.69&1.53&611.86\\
			\midrule
			dim$_{10}$&10&\textbf{1.04}&\textbf{337.71}&\textbf{10.83}&1.09&348.93&10.79&1.10&351.33&5.68&2.81&1258.89
			&6.73&1.84&1215.27\\
			\midrule
			dim$_{15}$&10&\textbf{1.03}&\textbf{781.53}&\textbf{11.10}&1.07&814.81&11.06&1.07&810.60&5.74&2.22
			&2485.81&7.47&1.38&2444.96\\
			\midrule
			R$_{15}$&10&\textbf{3.09}&19.92&9.32&3.24&\textbf{19.84}&\textbf{11.38}&3.36&20.16&7.81&5.33&28.95&8.57
			&4.68&28.53\\
			\midrule
			D$_{31}$&10&\textbf{6.43}&\textbf{124.58}&10.46&7.36&125.85&\textbf{11.17}&7.42&126.26&7.61&9.52&138.61
			&10.06&8.14&138.47\\
			\botrule
	\end{tabular}}

\end{table*}


\subsubsection{Experiment on the imbalanced-sets}\label{subsubsec:imbalanced-sets}

Figure \ref{fig:synthetic-p} ((a)-(d))
present the clustering results on the
imbalanced-sets for different $p$ with $N=3$.
It is seen that INCKPP${}_{sample}$ gets the best clustering
performance. 
Figure \ref{fig:synthetic-N} ((a)-(d)) present the min-SE against $N$ for all compared algorithms on the imbalanced-sets with $p=10$. It can be seen that for imbalance${}_2$,
imbalance${}_4$ and imbalance${}_6$ data sets, INCKPP${}_{sample}$ gets the best clustering results than the other compared
algorithms and achieves the global minimum first before $N=2,3,5$, respectively. For imbalance data set, INCKPP${}_{sample}$ and KPP
both achieve the global minimum before $N=20$, but INCKPP${}_{sample}$ gets a smaller min-SE for $N<20$. Further, from Table \ref{tab:synthetic} it is seen that
the average of the sum of errors (aver-SE) and the number of iterations ($\#iter$) of INCKPP${}_{sample}$
is generally much less than those of the compared algorithms. In conclusion, INCKPP${}_{sample}$
outperforms the other compared algorithms on the imbalanced-sets in the experiments.


\subsubsection{Experiments on the S-sets} \label{subsubsec:s-sets}

Figure \ref{fig:synthetic-p} ((e)-(h)) present the clustering results on the S-sets for different $p$ with $N=10$.
INCKPP${}_{sample}$ gets the smallest min-SE except
for the S${}_{4}$ data set,
for which KPP${}_{sample}$ gets a smaller min-SE than INCKPP${}_{sample}$ does when $p=8$.
Figure \ref{fig:synthetic-N} ((e)-(h)) present the min-SE against $N$ for all compared algorithms on the S-sets when $p=10$.
For S${}_{1}$ and S${}_{2}$ data sets , INCKPP${}_{sample}$
achieves the global minimum before $N=20$ and $N=30$, respectively, and gets a smaller min-SE than other compared algorithms. For
S${}_{3}$ and S${}_{4}$ data sets, non of the compared algorithms achieves the global minimum before $N=80$
but the min-SE value obtained by INCKPP${}_{sample}$ is smaller than others.
From Table \ref{tab:synthetic}, it is seen that INCKPP${}_{sample}$ obtains both the most number of repeats
within the same time and the smallest aver-SE and $\#iter$ values. Based on there results, it may be concluded that
the clustering performance of INCKPP${}_{sample}$ is better than that of the other compared
algorithms on the S-sets.


\subsubsection{Experiments on the Dim-sets}\label{subsubsec:Dim-sets}

Figure \ref{fig:synthetic-p} ((i)-(l)) show the clustering
performance on the Dim-sets
for different $p$ with $N=2$. It is seen that, for different $p$, INCKPP${}_{sample}$ gets the best
clustering performance on Dim-set except for
dim${}_{2}$ and dim${}_{10}$, for which the min-SE value obtained by KPP${}_{sample}$ is slightly smaller
than that obtained by INCKPP${}_{sample}$ when $p=14$.
Figure \ref{fig:synthetic-N} ((i)-(l)) present the min-SE against $N$ for all compared algorithms on the Dim-sets when $p=10$.
For dim${}_{2}$ and dim${}_6$ data sets, INCKPP${}_{sample}$ gets the
smaller min-SE value than other compared algorithms and achieves the global minimum
before $N=2$. For dim${}_{15}$ data set, both INCKPP${}_{sample}$ and KPP
attain the global minimizer before $N=3$ but the former gets a smaller min-SE value than the latter does
for $N<3$. For dim$_{10}$, both INCKPP${}_{sample}$ and KPP${}_{sample}$ attain the global minimizer
before $N=3$ but INCKPP${}_{sample}$ gets a smaller min-SE value compared with the latter.
Further, Table \ref{tab:synthetic} illustrates that the aver-SE value and the number $\#iter$ of iterations
of INCKPP${}_{sample}$ are both smaller than those of the other comparing methods. As a result,
INCKPP${}_{sample}$ outperforms the other compared algorithms for all the criteria on the Dim-sets.


\subsubsection{Experiments on the Shape-sets} \label{subsubsec:Shape-sets}

Figure \ref{fig:synthetic-p} ((m)-(n)) show the clustering performance on the Shape-sets for different $p$ with $N=10$.
INCKPP${}_{sample}$
get best result on Shape-sets except for R$_{15}$ data set when $p=8, 14$.
Figure \ref{fig:synthetic-N} ((m)-(n)) show the min-SE against $N$ for all compared algorithms on the Shape-sets for $p=10$.
It is found that INCKPP${}_{sample}$, KPP and
KPP${}_{sample}$ have a much better performance than FKM and FKM${}_{sample}$ on Shape-set.
For the R${}_{15}$ data set, INCKPP${}_{sample}$,
KPP${}_{sample}$ and KPP all achieve the global minimum at $N=20$, but INCKPP${}_{sample}$ and
KPP${}_{sample}$ obtained almost the same min-SE value before $N<20$ which is smaller than that
obtained by KPP. For the D${}_{31}$ data set, non of the compared
algorithms achieve the global minimum before $N=40$. However, INCKPP${}_{sample}$ gets a smaller min-SE
value for all $N$ compared with other four compared algorithms.
From Table \ref{tab:synthetic}, it is seen that INCKPP${}_{sample}$ obtains the smallest value of $\#iter$
on both D${}_{31}$ and R${}_{15}$ data sets. INCKPP${}_{sample}$ gets the smallest aver-SE
value on D${}_{31}$ data set and a slightly bigger aver-SE value than the smallest aver-SE value obtained
by KPP on R${}_{15}$ data set. Thus we may conclude that INCKPP${}_{sample}$ outperforms the other four algorithms
on D${}_{31}$ data set and has a similar performance on R${}_{15}$ data set with KPP${}_{sample}$.


\subsection{Experiments on the real datasets} \label{subsec:real}

In this subsection, we conduct experiments on seven real datasets to compare
our algorithms INCKPP and INCKPP${}_{sample}$ with INCKM, KPP, KPP${}_{sample}$, FKM and
FKM${}_{sample}$.

Table \ref{tab:Inckm-Inckpp-realworld} presents the experiment
results of INCKPP and INCKM on the real datasets: pendigits${}_3$, pendigits${}_5$,
pendigits${}_8$, pendigits, yeast, banknote, newthyroid.
The results illustrate that INCKPP achieves the smallest min-SE value (min-SE*) within the CPU time
used by INCKM on the seven data sets.

Hereinafter, we carry out extensive experiments to compare the fast version of
INCKPP (i.e., INCKPP${}_{sample}$) with KPP$_{sample}$, FKM$_{sample}$, KPP and FKM on real datasets.

\begin{table}[!thb]
	\centering
	\caption{The comparisons between INCKM and INCKPP on real datasets, where the meaning of the criteria is the same as
		in Table \ref{tab:Inckm-Inckpp-synthetic}}\label{tab:Inckm-Inckpp-realworld}
		\begin{tabular}{|c|c|c|c|c|c|c|}
			\toprule
			\multirow{2}{*}{Data sets}&\multicolumn{3}{c|}{INCKM}&\multicolumn{3}{c|}{INCKPP}\\
			\cmidrule{2-7}
			&$\lambda$&min-SE&\#it&min-SE&$\#repe$&\#it\\
			\midrule
			pendigits${}_3$&1.5&1673.30&\textbf{2.00}&\textbf{1466.98}&13.58&2.20\\
			\midrule
			pendigits${}_5$&1.8&2492.06&5.64&\textbf{2428.09}&17.06&\textbf{3.22}\\
			\midrule
			pendigits${}_8$&1.9&3664.18&4.09&\textbf{3619.82}&14.180&\textbf{2.86}\\
			\midrule
			pendigits&1.7&5125.03&\textbf{1.45}&\textbf{4968.93}&14.35&2.64\\
			\midrule
			yeast&1.7&293.22&1.64&\textbf{282.42}&11.85&\textbf{1.57}\\
			\midrule
			banknote&1.5&\textbf{406.68}&3.64&\textbf{406.68}&13.14&\textbf{3.38}\\
			\midrule
			newthyroid&2.2&\textbf{41.30}&\textbf{1.27}&\textbf{41.30}&11.81&2.06\\
			\botrule
		\end{tabular}

\end{table}

\begin{table*}[!thb]
	\centering
	\caption{The experiment results on real datasets with $p=10$, where the meaning of the criteria is
		the same as that in Table \ref{tab:synthetic}.}\label{tab:real}
\resizebox{\textwidth}{!}{
		\begin{tabular}{|c|c|c|c!{\vrule width 0.8pt}c|c|c!{\vrule width 0.8pt}c|c|c!{\vrule width 0.8pt}
				c|c|c!{\vrule width 0.8pt}c|c|c!{\vrule width 0.8pt}}
			\midrule
			\multirow{2}{*}{Data set}&\multicolumn{3}{c!{\vrule width 0.8pt}}{INCKPP${}_{sample}$}&
			\multicolumn{3}{c!{\vrule width 0.8pt}}{KPP}&\multicolumn{3}{c!{\vrule width 0.8pt}}{KPP${}_{sample}$}
			&\multicolumn{3}{c!{\vrule width 0.8pt}}{FKM}&\multicolumn{3}{c!{\vrule width 0.8pt}}{FKM${}_{sample}$}\\
			\cmidrule{2-16}
			&$\#re$& $\#it$& aver-SE& $\#re$& $\#it$&aver-SE& $\#re$& $\#it$&aver-SE& $\#re$& $\#it$&aver-SE&
			$\#re$& $\#it$&aver-SE\\
			\midrule
			pendigits&10&2.79&\textbf{5159.56}&7.71&4.15&5170.83&7.73&4.14&5170.10&7.54&4.55&5236.07&\textbf{10.58}&
			\textbf{2.70}&5220.36\\
			\midrule
			pendigits$_{8}$&\textbf{10}&\textbf{2.42}&\textbf{3854.31}&7.05&3.96&3964.72&7.16&3.96&3865.13&6.43&4.46&
			3972.75&\textbf{10}&2.48&3957.70\\
			\midrule
			pendigits$_{5}$&10&2.25&2525.51&7.20&3.61&2527.51&7.19&3.62&\textbf{2523.74}&6.61&3.98&2586.11&
			\textbf{10.48}&\textbf{2.13}&2580.08\\
			\midrule
			pendigits$_{3}$&\textbf{10}&1.63&\textbf{1565.10}&7.68&2.43&1573.32&7.59&2.41&1571.70&7.46&2.66&
			1649.26&9.96&\textbf{1.61}&1629.46\\
			\midrule
			yeast&10&2.59&292.80&9.48&2.94&\textbf{292.74}&9.80&2.96&293.05&10.83&3.02&296.26&\textbf{11.78}
			&\textbf{2.44}&295.48\\
			\midrule
			bankonte&10&1.99&\textbf{410.14}&8.22&2.68&410.88&8.28&2.66&411.08&7.71&2.79&411.73&\textbf{10.63}
			&\textbf{1.83}&411.14\\
			\midrule
			newthyroid&10&\textbf{2.18}&42.94&9.18&2.36&\textbf{42.84}&13.16&2.31&42.95&12.93&2.61&43.58
			&\textbf{14.06}&2.23&43.34\\
			\botrule			
	\end{tabular}}
	
\end{table*} 

\begin{figure*}[!thb]
	\centering
	\subfigure[pendigits]{\includegraphics[width=0.2\textwidth]{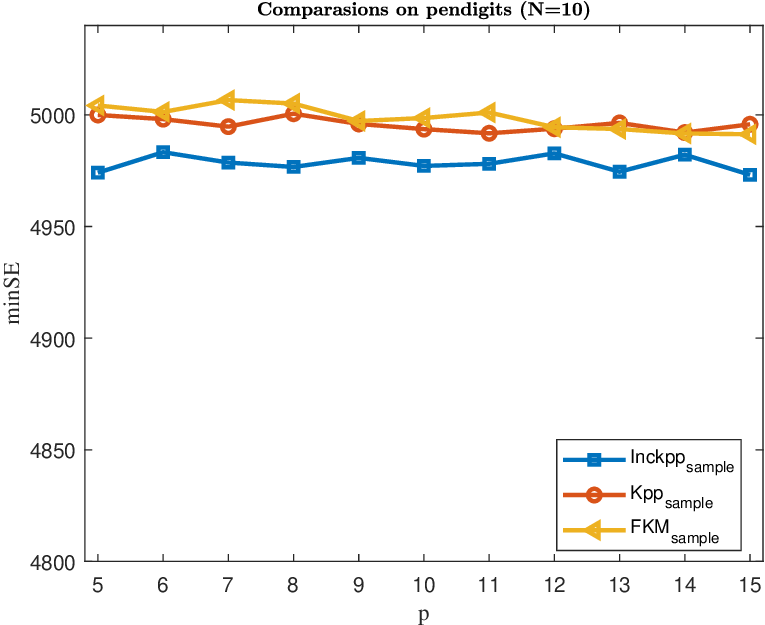}
	}
	\subfigure[pendigits${}_{8}$]{\includegraphics[width=0.2\textwidth]{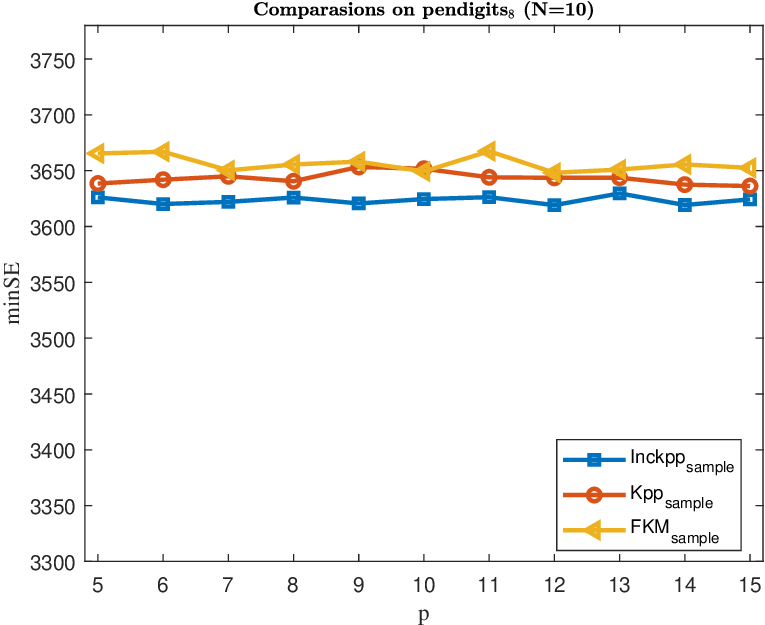}
	}
	\subfigure[pendigits${}_{5}$]{\includegraphics[width=0.2\textwidth]{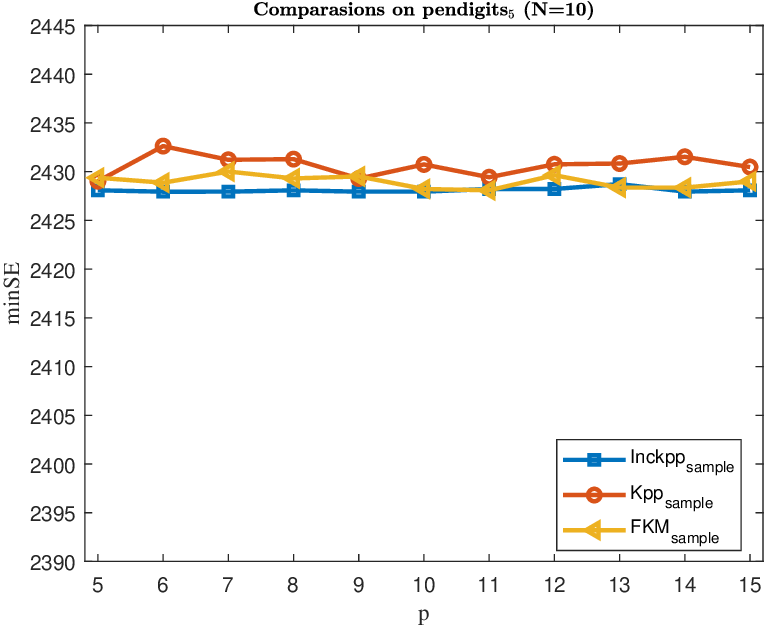}
	}
	\subfigure[pendigits${}_{3}$]{\includegraphics[width=0.2\textwidth]{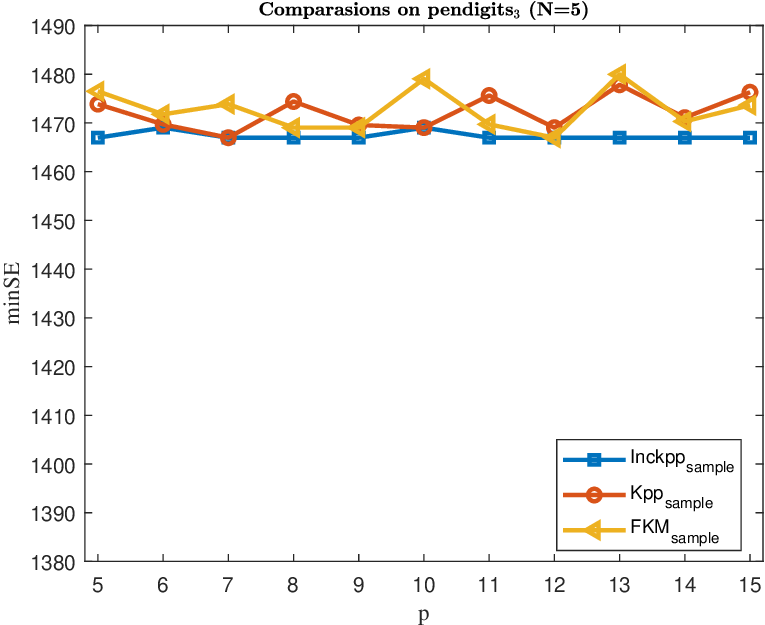}
	}
	\subfigure[yeast]{\includegraphics[width=0.2\textwidth]{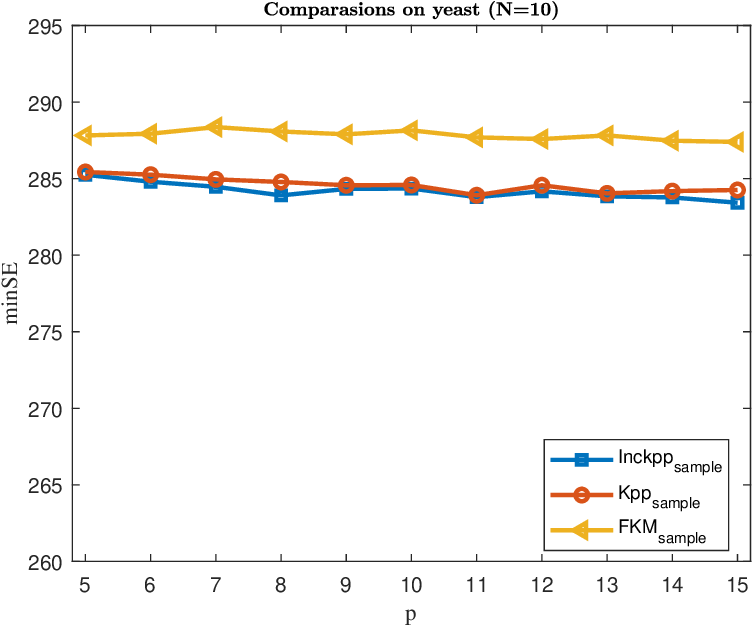}
	}
	\subfigure[banknote]{\includegraphics[width=0.2\textwidth]{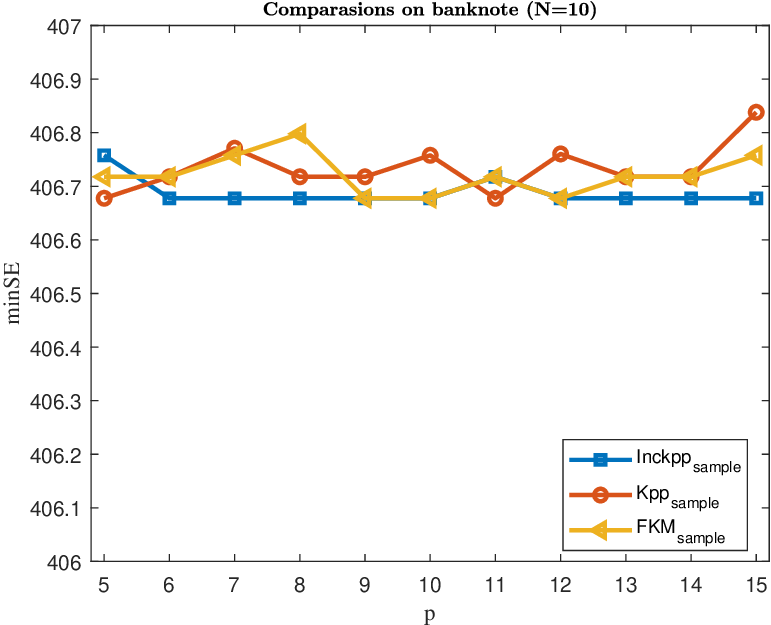}
	}
	\subfigure[newthyroid]{\includegraphics[width=0.2\textwidth]{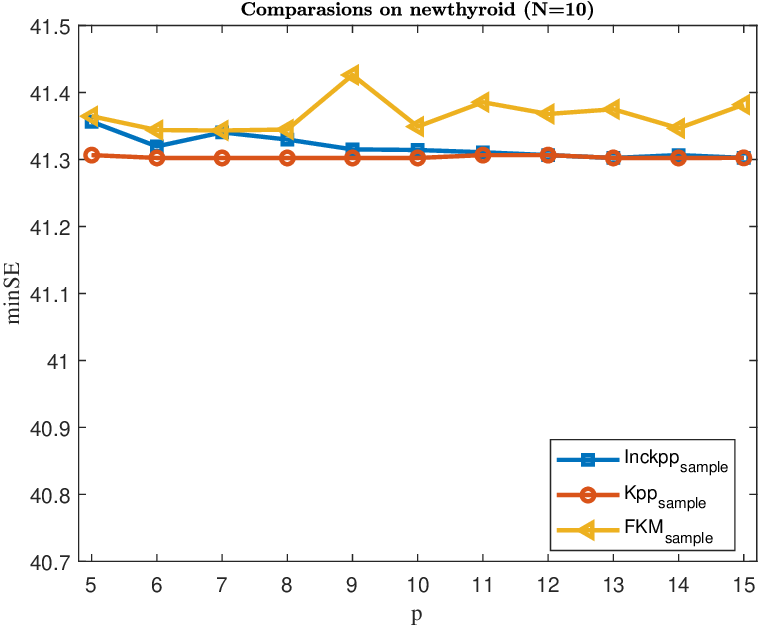}
	}
	\caption{The comparisons on the real datasets with different $p$,
	where $p$ and the values in the figures are the same as those in Figure \ref{fig:synthetic-p}.
	}\label{fig:real-p}
\end{figure*}

\begin{figure*}[!thb]
	\centering
	\subfigure[pendigits]{\includegraphics[width=0.2\textwidth]{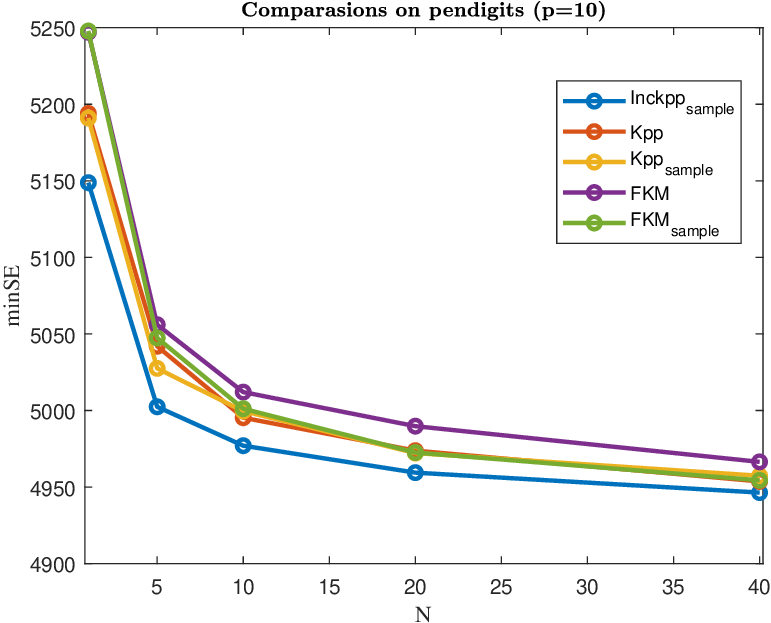}
	}
	\subfigure[pendigits${}_{8}$]{\includegraphics[width=0.2\textwidth]{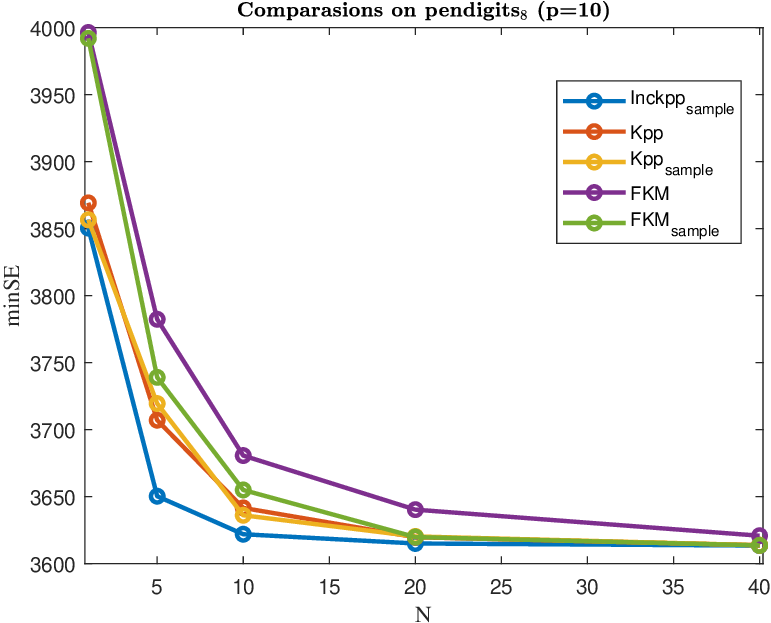}
	}
	\subfigure[pendigits${}_{5}$]{\includegraphics[width=0.2\textwidth]{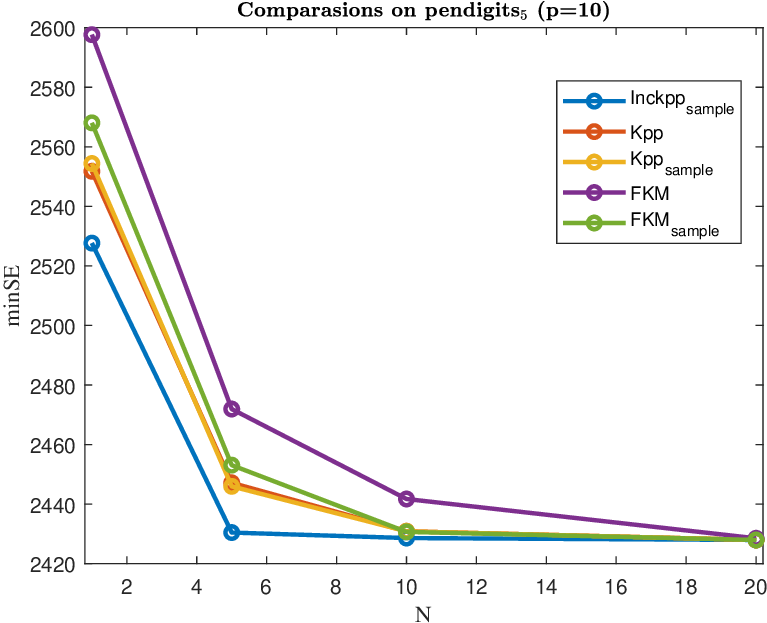}
	}
	\subfigure[pendigits${}_{3}$]{\includegraphics[width=0.2\textwidth]{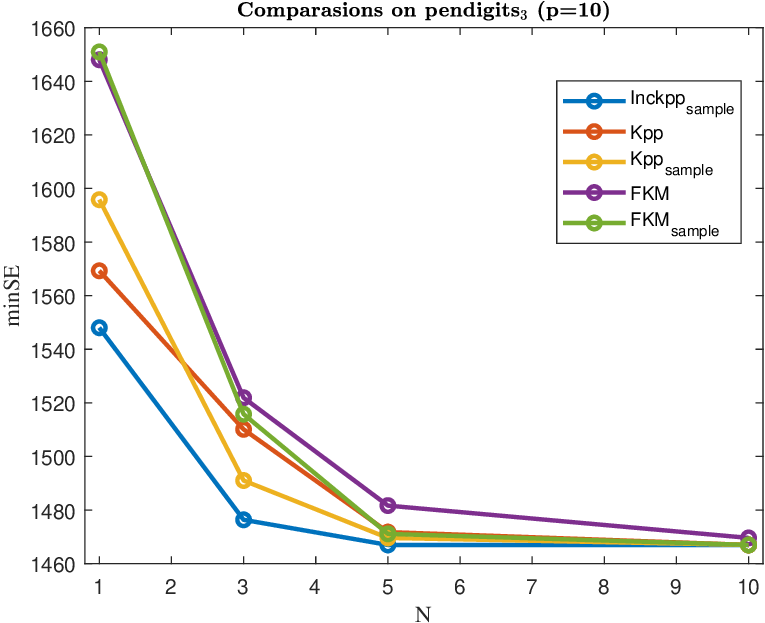}
	}
	\subfigure[yeast]{\includegraphics[width=0.2\textwidth]{p_yeast}
	}
	\subfigure[banknote]{\includegraphics[width=0.2\textwidth]{p_banknote}
	}
	\subfigure[newthyroid]{\includegraphics[width=0.2\textwidth]{p_newthyroid}
	}
	\caption{The comparisons on the real datasets for different $N$, where $N$ and the values in the figures are the same as those in Figure \ref{fig:synthetic-N}.
	}\label{fig:real-N}
\end{figure*}


\subsubsection{Experiments on the Handwritten digits-sets}\label{subsubsec:handwritten-sets}

Figure \ref{fig:real-p} ((a)-(d)) show the clustering performance on the handwritten digits-sets for different $p$ with $N=10$.
The results illustrate that
INCKPP${}_{sample}$ gets the best clustering performance with the smallest min-SE on the handwritten digits-sets except for the pendigits${}_5$
data set, for which FKM${}_{sample}$ gets a smaller min-SE than INCKPP$_{sample}$
does when p = 11 and 13.
Figure \ref{fig:real-N} ((a)-(d)) present the min-SE against $N$ for all compared algorithms on the handwritten digits-sets for different $N$ with $p=10$.
For pendigits${}_8$, pendigits${}_5$ and
pendigits${}_3$ data sets, INCKPP${}_{sample}$ gets a smaller min-SE value compared with the other four algorithms
and first achieves the global minimum at $N=20,10,5$, respectively.
For pendigits data set, all five compared algorithms do not achieve
the global minimum before $N=40$ but INCKPP${}_{sample}$ gets a much smaller min-SE value for each $N$
compared with the other four algorithms. 
Table \ref{tab:real} shows that INCKPP${}_{sample}$
obtains the smallest aver-SE value on pendigits, pendigits${}_8$ and pendigits${}_3$ data sets and a slightly
larger aver-SE value than that obtained by KPP${}_{sample}$ on pendigits${}_{5}$ data set.
In addition, FKM${}_{sample}$ obtains the smallest $\#repe$ value on pendigits, pendigits${}_5$ and
pendigits${}_3$ data sets and a lightly larger $\#repe$ value than that obtained by INCKPP${}_{sample}$ on
pendigits${}_{8}$ data set. In conclusion, INCKPP${}_{sample}$ outperforms the other compared algorithms on the handwritten digits-sets.


\subsubsection{Experiments on other real datasets}\label{subsubsec:real2}

Figure \ref{fig:real-p} ((e)-(g)) show clustering performance on the yeast, banknote and newthyroid data sets for different $p$ with $N=10$.
For yeast data set, it is seen that INCKPP${}_{sample}$ get
the smallest min-SE value.
From Fig \ref{fig:real-p}(f), it is observed that INCKPP${}_{sample}$ gets the
the smallest min-SE value on the banknote data set among the three algorithms except for $p=5$
where both KPP${}_{sample}$ and FKM${}_{sample}$ achieve a smaller min-SE value
than that obtained by INCKPP${}_{sample}$ and for $p=11$ where INCKPP${}_{sample}$
and FKM${}_{sample}$ get the same min-SE value which is bigger than that got by KPP${}_{sample}$,
whilst from Fig \ref{fig:real-p}(g), it is found that, on the newthyroid data set,
KPP${}_{sample}$ gets the smallest min-SE value and FKM${}_{sample}$ gets the largest min-SE value
among the three compared algorithms for $p<10$, but for $p\ge 10$
INCKPP${}_{sample}$ and KPP${}_{sample}$ almost get the same min-SE value which is much smaller than
that obtained by FKM${}_{sample}$. It is then concluded from the above discussions that INCKPP${}_{sample}$
achieves an overall better clustering performance than KPP${}_{sample}$ and FKM${}_{sample}$ have
on the three data sets.
Figure \ref{fig:real-N} ((e)-(g)) presents the min-SE against $N$ of INCKPP${}_{sample}$,
KPP, KPP${}_{sample}$, FKM and FKM${}_{sample}$ onthe yeast, banknote and newthyroid data sets for $p=10$.
Figure \ref{fig:real-N} (e) shows that INCKPP${}_{sample}$ gets the smallest min-SE value
on the yeast data set for all values of $N$ among the five comparing algorithms.
It is seen from Figure \ref{fig:real-N} (f) that for the banknote data set INCKPP${}_{sample}$
and FKM${}_{sample}$ achieve the global minimum at $N=5$ and $N=7$, respectively. 
From Figure \ref{fig:real-N} (g), it is seen that for the newthyroid data set 
KPP${}_{sample}$ has the best performance among the five comparing algorithms 
and is the only one which achieves the global minimum at $N=5$ with the smallest min-SE value among 
those obtained by the five comparing algorithms,
whilst the min-SE value obtained by INCKPP${}_{sample}$ decreases faster than those obtained by KPP, 
FKM and FKM${}_{sample}$ do with $N$ increasing and is the second smallest at $N=5$. 
Combine the results in the Table \ref{tab:real},, the clustering performance of INCKPP${}_{sample}$ is better than that of
the other four comparing algorithms on the yeast and banknote datasets, but
KPP${}_{sample}$ outperforms the other four comparing algorithms on the newthyroid data set.


\section{Conclusions} \label{sec:conclusion}

In this paper, we proposed a novel incremental $k$-medoids algorithm (named INCKPP).
INCKPP addressed the parameter selection issue in the improved $k$-medoids algorithm \cite{yu2018improved} and thus 
can dynamically increase the number of clusters from $2$ to $k$ through a non-parametric and 
stochastic $k$-means$++$ search procedure. INCKPP outperforms than the improved $k$-medoids algorithm \cite{yu2018improved} and can also deal with imbalanced datasets very well.
We further proposed a modified INCKPP algorithm (named INCKPP${}_{sample}$) which is fast and 
improves the computational efficiency of INCKPP without affecting its clustering performance.
Extensive experiments results on synthetic and real datasets including imbalanced-sets 
have been conducted in comparison with three state-of-the-art and the most commonly used algorithms: the improved $k$-medoids 
algorithm (INCKM) \cite{yu2018improved}, the simple and fast $k$-medoids algorithm (FKM) \cite{park2009simple} and the $k$-means$++$ algorithm (KPP) \cite{arthur2007k}. The experimental results on each data set consistently show that the proposed algorithm INCKPP$_{sample}$ has better performance than the compared algorithms, and it obtains better solutions in the same running time (corresponding to smaller value of loss function SE), and the INCKPP and INCKPP$_{sample}$ algorithm convergence requires fewer iterations, so the time required for a single run of the algorithm is shorter.
In the past few years, generative adversarial networks \cite{creswell2018generative,goodfellow2020generative} have been proposed and widely used in many fields such as intelligent transport systems \cite{zhou2022gan} and intelligent trust management in 6G wireless networks \cite{yang2022generative} and have achieved remarkable success in many unsupervised learning tasks. As an important class of unsupervised learning methods, the clustering algorithms are considered to be combined with generative adversarial networks in our future work.
\bibliography{Ikpp_ref}

\end{document}